\documentclass[letterpaper]{article} 
\usepackage{aaai24}  
\usepackage{times}  
\usepackage{helvet}  
\usepackage{courier}  
\usepackage[hyphens]{url}  
\usepackage{graphicx} 
\usepackage{natbib}  
\usepackage{caption} 
\usepackage{algorithm}
\usepackage{algorithmic}
\usepackage{xcolor}         
\usepackage{soul}

\usepackage{amsthm}
\usepackage{amssymb}
\usepackage{mathtools}
\usepackage{graphics}
\newtheorem{theorem}{Theorem}[section]

\newtheorem{lemma}[theorem]{Lemma}

\usepackage{courier}

\usepackage{bm}
\usepackage{makecell}
\usepackage{multirow}
\usepackage{lipsum,booktabs}
\usepackage{subfigure}
\usepackage{parskip}
\usepackage{subcaption}

\usepackage{graphicx}
\usepackage{comment}

\usepackage{newfloat}
\usepackage{listings}
\DeclareCaptionStyle{ruled}{labelfont=normalfont,labelsep=colon,strut=off} 
\floatstyle{ruled}
\newfloat{listing}{tb}{lst}{}
\floatname{listing}{Listing}
\title{ACT: Empowering Decision Transformer with Dynamic Programming via Advantage Conditioning}
\author {
    Chen-Xiao Gao, 
    Chenyang Wu, 
    Mingjun Cao, 
    Rui Kong, 
    Zongzhang Zhang\thanks{Zongzhang Zhang is the corresponding author.},
    Yang Yu
}
\affiliations {
    National Key Laboratory for Novel Software Technology, Nanjing University, China\\
    School of Artificial Intelligence, Nanjing University, China\\
    \{gaocx, wucy, caomj, kongr\}@lamda.nju.edu.cn,
    \{zzzhang, yuy\}@nju.edu.cn
}

\usepackage{bibentry}

\begin{document}

\maketitle

\begin{abstract}
Decision Transformer~(DT), which employs expressive sequence modeling techniques to perform action generation, has emerged as a promising approach to offline policy optimization. However, DT generates actions conditioned on a desired future return, which is known to bear some weaknesses such as the susceptibility to environmental stochasticity. To overcome DT's weaknesses, we propose to empower DT with dynamic programming. Our method comprises three steps. First, we employ in-sample value iteration to obtain approximated value functions, which involves dynamic programming over the MDP structure. Second, we evaluate action quality in context with estimated advantages. We introduce two types of advantage estimators, IAE and GAE, which are suitable for different tasks. Third, we train an Advantage-Conditioned Transformer~(ACT) to generate actions conditioned on the estimated advantages. Finally, during testing, ACT generates actions conditioned on a desired advantage. Our evaluation results validate that, by leveraging the power of dynamic programming, ACT demonstrates effective trajectory stitching and robust action generation in spite of the environmental stochasticity, outperforming baseline methods across various benchmarks. Additionally, we conduct an in-depth analysis of ACT's various design choices through ablation studies. Our code is available at \url{https://github.com/LAMDA-RL/ACT}.
\end{abstract}

\section{Introduction}
Reinforcement Learning~(RL) optimizes policies by interacting with the environment often for millions of steps~\cite{sac,liu2018}. Such enormous sample complexity prohibits RL from real-world applications~\cite{wu2023} such as robotics and healthcare. As an alternative, offline RL optimizes policies with a pre-collected dataset~\cite{levine_survey} and has gained increasing attention in recent years for its potential in real-life scenarios~\cite{robotics1,healthcare1,zhou2023,zhou2024}.

Building upon online RL, a lot of algorithms address offline policy optimization following the spirit of dynamic programming~\cite{levine_survey}, i.e. leveraging the structure of Markov Decision Process~(MDP) and employing Bellman update to derive value estimates for subsequent policy optimization~\cite{bcq,iql,onestep_rl}.

In the meantime, the past few years have witnessed huge success in applying sequence modeling to natural language processing~\cite{attention_is_all_you_need, gpt3}. In light of the similarity between language sequences and RL trajectories, a lot of works have explored the idea of modeling RL trajectories using sequence modeling approaches~\cite{wen2023}. For example, Decision Transformer~(DT)~\cite{dt} models offline trajectories extended with the sum of the future rewards along the trajectory, namely the return-to-go~(RTG). RTG characterizes the quality of the subsequent trajectory, and DT learns to predict future actions given RTG. During testing, we deliberately provide DT with a high RTG to generate an above-average action. This approach, known as \textit{conditional sequence generation}, has achieved remarkable success in offline policy optimization.

However, we note that RTG-conditioned generation is defective. Firstly, RTG is a hindsight indicator of a complete trajectory. It cannot be computed if we only have incomplete trajectory segments. Besides, we cannot ascertain the most suitable target RTG during testing, which poses a risk of choosing an inappropriate one and degenerating the performance. Secondly, conditioning on RTG fails to leverage the inherent structure of MDPs and cannot stitch trajectory as classical offline RL methods do. Therefore, DT cannot achieve much better performance than RL methods. Lastly, when the environment is stochastic, DT tends to exploit the occasional high return in the offline dataset mistakenly, which negatively impacts its performance. 

In this paper, we present Advantage-Conditioned Transformer~(ACT), a method that empowers the traditional DT with dynamic programming to effectively address the above-mentioned challenges. To tackle the limitations of conditioning on RTG, we introduce advantages as replacements. To obtain the advantage, we first approximate value functions with separate neural networks. Afterward, we introduce two types of advantage estimations, namely IAE and GAE, which exhibit different characteristics and are proven suited for different types of tasks empirically. We evaluate ACT in various benchmarks including those with stochastic dynamics and delayed rewards. The results show that, with suitable advantage estimation, ACT significantly outperforms existing sequence modeling techniques and achieves performance on par with state-of-the-art offline RL methods. We also carried out extensive ablation studies to highlight the effectiveness of the design choices of ACT. Overall, our empirical results affirm the efficacy of the proposed method in addressing the limitations of DT and showcase its potential in applications.

\section{Preliminaries}
\subsection{Offline Reinforcement Learning}
A Markov Decision Process~(MDP) can be denoted by a six-tuple $\langle \mathcal{S}, \mathcal{A}, T, \mathcal{R}, \rho_0, \gamma \rangle$, where $\mathcal{S}$ stands for the state space, $\mathcal{A}$ is the action space, $T$ is the transition function, $\mathcal{R}$ is the reward function, $\rho_0$ is the initial distribution of states, and $\gamma\in[0, 1)$ denotes the discount factor of future rewards. For states $s,s'\in\mathcal{S}$ and $a\in\mathcal{A}$, $T(s'|s, a)$ specifies the transition probability of arriving at state $s'$ after taking action $a$ at state $s$, and $\mathcal{R}(s, a)$ gives the immediate reward of taking action $a$ at state $s$. A policy $\pi$ is a function mapping states to action distributions, and $\pi(a | s)$ is the probability of taking action $a$ at state $s$. The policy's quality is measured by the expected discounted return $J(\pi)=\mathbb{E}_\pi\left[\sum_{t=0}^\infty\gamma^{t}r_t\right]$, where $r_t=\mathcal{R}(s_t,a_t)$ is the reward obtained at time $t$, $s_t, a_t$ are the state and action at time $t$, respectively, and the expectation is taken w.r.t. the stochastic interaction of the policy $\pi$ and the MDP environment. In offline RL, an agent is expected to learn a policy maximizing the expected discounted return with a static dataset $\mathcal{D}$. Typically, $\mathcal{D} = \{\tau_k\}_{k=1}^{K}$ is composed of $K$ trajectories $\tau_k$ collected by a behavior policy $\beta$, where $\tau_k=\{s^k_t, a^k_t, r^k_t\}_{t=0}^{N_k-1}$, $s^k_t$, $a^k_t$, and $r^k_t$ denote state, action, and reward at timestep $t$ of the $k$-th trajectory, respectively, and $N_k$ is the trajectory length of the $k$-th trajectory.

\subsection{Dynamic Programming in RL}\label{sec2.2}
A commonly used categorization of deep RL algorithms is by their optimization paradigm, either \textit{policy gradient} or \textit{approximate dynamic programming}~\cite{levine_survey}. The former derives the gradient of RL objectives w.r.t. the policy via the policy gradient theorem~\cite{policy_gradient} and improves the policy via gradient ascent. The latter derives the optimal policy exploiting the value function. The value functions of a policy $\pi$ are defined recursively:
$$
    \begin{aligned}V^\pi(s)&=\mathbb{E}_{a\sim\pi(\cdot|s)}\left[Q^\pi(s,a)\right],\\
    Q^\pi(s, a)&=\mathcal{R}(s,a)+\gamma\mathbb{E}_{s'\sim T(\cdot|s,a)}\left[V^\pi(s')\right].
    \end{aligned}
$$
Solving the value function with dynamic programming, we can derive an improved policy by increasing the probability of selecting actions with positive advantages, where the advantage of an action $a$ at state $s$ is $A^{\pi}(s,a)=Q^{\pi}(s,a)-V^{\pi}(s)$. We can get the optimal policy by repeating this process, which is called policy iteration. Alternatively, we can use the value iteration and directly solve the optimal value functions $V^*$ and $Q^*$ with dynamic programming, where
$$
    \begin{aligned}
    V^*(s)&=\max\nolimits_a Q^*(s,a),\\
    Q^*(s, a)&=\mathcal{R}(s,a)+\gamma\mathbb{E}_{s'\sim T(\cdot|s,a)}\left[V^*(s')\right].
    \end{aligned}
$$
The optimal policy $\pi^*$ is then obtained by greedily selecting the action $a$ with the optimal value $Q^*(s,a)$ at all $s\in\mathcal{S}$. 

Compared with policy gradient methods, dynamic programming leverages the structure of MDP, gaining advantages in efficiency and performance empirically. 

\subsection{Decision Transformer}
One of the pioneering works of solving control tasks with high-capacity sequence modeling networks is the Decision Transformer~(DT)~\cite{dt}. For each trajectory $\tau\in\mathcal{D}$, DT first computes the RTG $\hat{R}_t$ for each timestep $t$ by summing up the rewards along the future trajectory, $\hat{R}_t=\sum_{t'=t}^{N-1}r_{t'}$, where $N$ is the trajectory length. Later, DT fits a GPT-2 model~\cite{gpt2} on the offline trajectories augmented with RTGs, $\tau^{\mathrm{RTG}} = \{\hat{R}_t, s_t, a_t\}_{t=0}^{N-1}$.

During testing, DT first specifies the desired target return as $\hat{R}_0$, and executes the predicted action $\hat{a}_0$ given the history $(\hat{R}_0, s_0)$. After observing the new state $s_{1}$ and reward $r_0$, DT sets $\hat{R}_1=\hat{R}_0 - r_0$ and continues to predict the next action given the updated history $(\hat R_0,s_0,\hat R_1,s_1)$. This process continues until the end of the episode. 

\section{Defects with RTG Conditioning}\label{sec3}
Existing DT algorithms are commonly implemented by RTG conditioning. Although RTG conditioning is relatively simple to implement, this naive choice has several defects. 

\begin{figure}[htbp]
    \centering
    \includegraphics[width=0.95\linewidth]{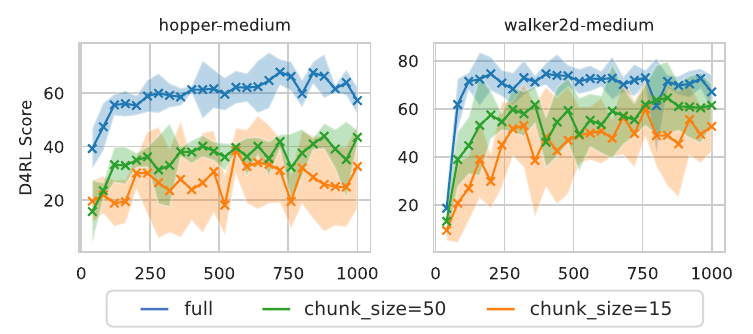}
    \caption{The performance of DT when the trajectory is divided into chunks. \textit{Chunk\_size} denotes the granularity of division, and \textit{full} means the complete trajectory is preserved. }
    \label{fig:dt_chunk}
\end{figure}

\textbf{Dependency on future trajectory. }
The calculation of RTG involves summing up the rewards in the future trajectory. However, in real-world scenarios, the sampling process is often susceptible to unexpected events, making it challenging to gather complete and continuous trajectories. Under such circumstances, the collected offline dataset may consist of broken snippets of trajectories or even independent transition tuples $(s,a,r,s')$, making it difficult to calculate RTG since the future is missing. As illustrated in Figure~\ref{fig:dt_chunk}, if we divide the trajectories into chunks, DT's performance will degenerate severely. 

\textbf{Inability to perform stitching. }
A well-established consensus on why offline RL methods outperform imitation learning is that they can perform \textit{trajectory stitching}~\cite{why_offlinerl_is_better}, i.e. concatenating multiple snippets of trajectories together to give better performance. However, as pointed out in recent works~\cite{qdt,edt}, DT lacks stitching ability, because the RTG is only associated with the current trajectory. Although conditioning on high RTGs might skew the generated action towards better ones, the performance of DT still falls behind traditional offline RL methods. 

\textbf{Failure in stochastic environments. }
The last failure mode of DT relates to its propensity to make optimistic decisions in stochastic settings, which arises from the intrinsic feature of RTG conditioning to imitate previous achievements indiscriminately while disregarding the randomness.

This phenomenon is also perceived as the conflation of the effects of the policy and the world model~\cite{doc}, meaning that RTG fails to distinguish the controllable parts (the policy) from the uncontrollable parts (the dynamics) and, as a result, biases the action generation with blind optimism about the dynamics.

\begin{figure*}[htbp]
    \centering
    \includegraphics[width=0.75\linewidth]{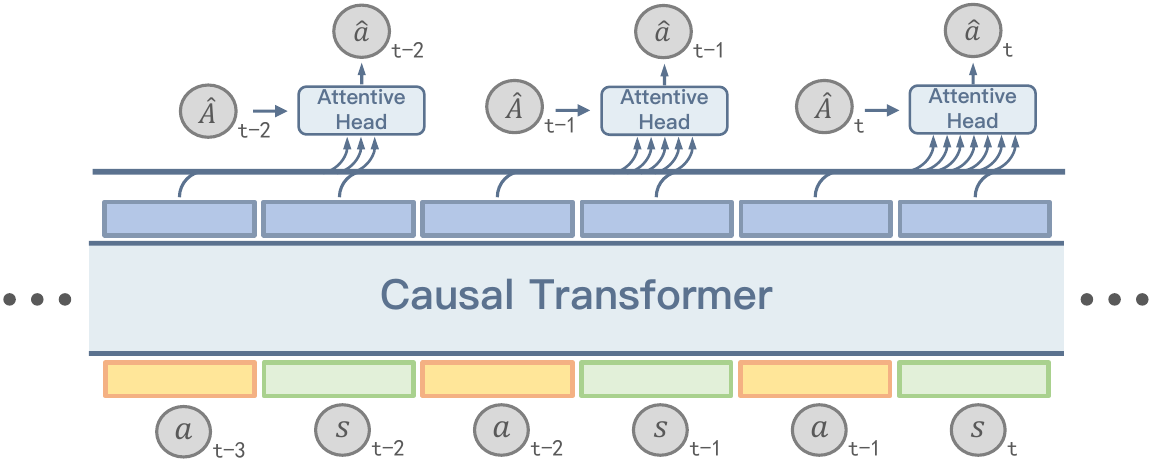}
    \caption{The encoder-decoder architecture of ACT. The encoder encodes the historical state-action sequence into a continuous representation. The attentive head of the decoder queries historical representation with the advantage and predicts an action.}
    \label{fig:arch}
\end{figure*}

\section{Advantage-Conditioned Transformer}
The previous discussion motivates us to devise a better alternative in place of RTG. Observing that dynamic programming handles these challenges effectively, we decide to exploit it for conditional generation. 

Specifically, we approximate the value function via approximate dynamic programming~(Section~\ref{sec4.1}) and estimate advantages with the learned value functions.
After labeling the dataset with advantages~(Section~\ref{sec4.2}), we fit an encoder-decoder transformer on the advantage-augmented dataset via the action reconstruction loss~(Section~\ref{sec4.3}).
We additionally train a predictor $c_\phi$ to estimate the maximal advantage in samples and condition the action generation on the estimated maximum advantage during testing.
The complete algorithm is presented in Algorithm~\ref{algorithm}.

\subsection{Value Function Approximation}\label{sec4.1}
We use two parameterized functions $Q_\theta$ and $V_\psi$ to approximate the state-action value function and the state value function, respectively. With the offline dataset, we iteratively update the parameters via stochastic gradient descent,
\begin{equation}\label{eq:dp}
\small{
\begin{aligned}
    &\theta\gets \theta-\frac{\eta}{M}\sum_{(s,a,r,s')\in\mathcal{B}}\nabla_\theta\left(r + \gamma V_{\bar{\psi}}(s')-Q_\theta(s, a)\right)^2,\\
    &\psi\gets \psi-\frac{\eta}{M}\sum_{(s,a,r,s')\in \mathcal{B}}\nabla_\psi\mathcal{L}_{\sigma_1}\left(r+\gamma V_{\bar{\psi}}(s') - V_\psi(s)\right), 
\end{aligned}
}
\end{equation}

where $\mathcal{B}=\{(s_i,a_i,r_i,s'_i)\}_{i=1}^M$ denotes a mini-batch of transition tuples sampled uniformly from the dataset $\mathcal{D}$, $M$ is the batch size, $\eta$ is the learning rate, $V_{\bar{\psi}}$ is the target network, $\sigma_1$ is a hyper-parameter, and $\mathcal{L}_{\sigma_1}(u) = |\sigma_1 - \mathbb{I}(u<0)|u^2$ is the $\sigma_1$-expectile regression loss.

Equation~\eqref{eq:dp} performs in-sample value iteration as outlined by~\citet{iql} which does not query out-of-distribution actions as classical value iteration does. When $\sigma_1$ equals $0.5$, it is conducting on-policy value evaluation, producing an approximation of $Q^\beta$ and $V^\beta$. With infinite samples and $\sigma_1 \to 1$, it produces an approximation of the in-sample optimal value functions: 
$$
    \begin{aligned}
    V_\beta^*(s)&=\max\nolimits_{\beta(a|s)>0} Q_\beta^*(s,a),\\
    Q_\beta^*(s, a)&=\mathcal{R}(s,a)+\gamma\mathbb{E}_{s'\sim T(\cdot|s,a)}\left[V_\beta^*(s')\right].
    \end{aligned}
$$
Thus, this learning procedure smoothly interpolates between approximating on-policy value functions and the in-sample optimal value functions by varying the value of $\sigma_1$. 

Upon the convergence of the training, we freeze $\theta$ and $\psi$ and do not make further updates to them.

\subsection{Advantage Labeling}\label{sec4.2}
In the next step, we utilize the pre-trained value function approximation to tag the offline data with their advantage. The most straightforward advantage estimator $\hat{A}_{\text{IAE}}$ is simply the difference between $Q_\theta$ and $V_\psi$:
$$
    \begin{aligned}
    \hat{A}^{\text{IAE}}_t = Q_\theta(s_t, a_t) - V_\psi(s_t). 
    \end{aligned}
$$
We term this \textit{Immediate Advantage Estimation}~(IAE). Since $Q_\theta$ averages all subsequent random variations, IAE produces an estimator robust to transitional stochasticity.

However, IAE issues a demand on estimating the state-action value function, which, compared to the state value function, has higher dimensional input and is more difficult to estimate. Besides, solely depending on Temporal Difference~(TD) error to optimize the value functions is known to fail in sparse-reward or goal-oriented tasks~\cite{qdt, dwsl}. Thus, we propose to use the \textit{Generalized Advantage Estimation}~(GAE)~\cite{gae} as an alternative:
$$
    \hat{A}^{\text{GAE}(\lambda)}_t =(1-\lambda)\sum_{l=1}^\infty \lambda^{l-1}\hat{A}^{(l)}_t,
$$
where $\hat{A}^{(l)}_t=-V_\psi(s_t)+\sum_{i=0}^{l-1} \gamma^{i}r_{t+i}+\gamma^lV_\psi(s_{t+i+1})$ denotes the $l$-step advantage along the given trajectory. Note that GAE can be extended to the finite horizon case, though we present the infinite horizon formulation here for simplicity. The hyper-parameter $\lambda\in[0,1]$ interpolates between TD estimation and Monte-Carlo estimation, controlling the bias-variance trade-off. With $\lambda$ set to $1$, we have $\hat{A}^{\mathrm{GAE}(1)}_t=\sum_{i=0}^\infty\gamma^i r_{t+i} - V_\psi(s_t)$, which is functionally equivalent to the discounted RTG, since the constant $-V_\psi(s_t)$ does not count when comparing actions.

After deciding on the form of advantage estimator, we label the offline dataset with their advantages, and construct the advantage-augmented dataset $\mathcal{D}^{\mathrm{A}}=\{\tau_k^{\mathrm{A}}\}_{k=1}^{M}$, where $\tau^{\mathrm{A}}_k = \{\hat{A}_t^k, s_t^k, a_t^k\}_{t=0}^{N_k-1}$, and $\hat{A}_t^k$ is the estimated advantage at the $t$-th timestep of the $k$-th trajectory by either IAE or GAE.

\subsection{Advantage-Conditioned Sequence Modeling}\label{sec4.3}
The original DT straightforwardly applies the GPT-2 architecture to RL sequence modeling, offering convenience but falling short of optimization for decision-making tasks.
In our experiments, we observed that modeling using GPT-2 architecture causes overfitting and unstable generation during testing, aligning well with recent research findings~\cite{unimask}.
In light of this, we propose an encoder-decoder transformer architecture illustrated in Figure~\ref{fig:arch}.

In our architecture, the encoder only accepts the state-action sequence as input and applies a causal attention mechanism~\cite{attention_is_all_you_need} to mask out future inputs. It transforms the input sequence $(s_{<t}, a_{<t}, s_t)$ to a representation sequence, where $s_{<t}$ and $a_{<t}$ denote the states and actions before timestep $t$, respectively. 
With cross attention, the decoder queries historical representation with the estimated advantage $\hat A_t$ of the current timestep and predicts an action $\hat a_t$.
The complete function is represented as $\hat{a}_t=\texttt{ACT}(s_{<t}, a_{<t}, s_t, \hat{A}_t)$. 
We found that the separation of the state-action sequence and the advantage benefits the overall performance. Moreover, such separation leaves room for self-supervised pre-training. We may pre-train the encoder using a large unlabelled dataset by techniques similar to Masked Language Modeling~(MLM), reinitialize the decoder, and fine-tune the transformer to accomplish control tasks~\cite{smart, unimask, mtm}. 

Another change we made is using the sinusoidal positional encoding in place of learnable positional embedding because we found the latter one sometimes causes instability, which echoes the finding by~\citet{onlinedt}. Other details are deferred to Appendix~B.1\footnote{\url{https://www.lamda.nju.edu.cn/gaocx/AAAI24-supp.pdf}} due to the limitation of space. 

The loss for training ACT is the action reconstruction loss,
\begin{equation}\label{eq:act_objective}
    \mathcal{L}_{\mathrm{ACT}}=\mathbb{E}_{\tau^{\mathrm{A}}\sim\mathcal{D}^{\mathrm{A}}}\left[\sum_{t}\Big(a_t-\texttt{ACT}(s_{<t}, a_{<t}, s_t, \hat{A}_t)\Big)^2\right],
\end{equation}
where $\tau^{\mathrm{A}}\sim\mathcal{D}^{\mathrm{A}}$ means sampling $\tau^{\mathrm{A}}$ uniformly from $\mathcal{D}^{\mathrm{A}}$. 

Finally, similar to DT, we need to specify the target advantage for each state during testing. We additionally train a predictor network $c_\phi$ with the expectile regression loss,
\begin{equation}\label{eq:condition_objective}
    \begin{aligned}
    \mathcal{L}_{\phi}=\mathbb{E}_{(\hat{A}_t, s_t)\sim \mathcal{D}^{\mathrm{A}}}\left[\mathcal{L}_{\sigma_2}\left(\hat{A}_t - c_\phi(s_t)\right)\right],
    \end{aligned}
\end{equation}
where $(\hat{A}_t, s_t)\sim \mathcal{D}^{\mathrm{A}}$ denotes that $(\hat{A}_t, s_t)$ is sampled uniformly from $\mathcal{D}^{\mathrm{A}}$. In the limit of $\sigma_2\to1$, $c_\phi$ learns to predict the maximal in-sample advantage 
 given state $s$, which fully exploits ACT's potential to generate a good action.

\subsection{Why Advantage Conditioning Offers Benefits?}\label{sec4.4:benefits}

Recall the Performance Difference Lemma \cite{policy_difference_lemma}, the performance difference between two policies $\pi$ and $\pi'$ can be expressed as the expected advantage w.r.t. $\pi'$ on the state-action distribution induced by $\pi$, $\eta(\pi)-\eta(\pi')=\mathbb{E}_{\tau\sim\pi}\left[\sum_{t=0}^{\infty}\gamma^t A^{\pi'}(s_t, a_t)\right]$. Assume that the dataset is labeled using the advantage function of $\pi'$ and our learners are exempt from sampling and approximation errors, then when conditioned on a desired advantage $A$, ACT will generate an action $a$ satisfying $A^{\pi'}(s, a)=A$. By auto-regressively generating a trajectory and assigning ACT a positive target advantage at each timestep, ideally, all actions in the sequence possess positive advantages. Taking the expectation, we have $\mathbb{E}_{\tau\sim p_\pi}[\sum \gamma^t A_t^{\pi'}]>0$ indicating that the induced ACT policy improves over the behavior policy $\pi'$ by the lemma.
As discussed previously, with $\sigma_1=0.5$, the learned value function approximates the value function of the behavior policy $\beta$, i.e., $\pi'=\beta$. If $\sigma > 0.5$, then the estimated advantage corresponds to a policy $\pi'$ that is already improved over $\beta$.

Since RTGs are Monte-Carlo value estimates for the value of the behavior policy $\beta$, DT is restricted to improve on top of $\beta$, which limits the potential for improvement compared with ACT. Moreover, RTGs suffer from high variance, often assigning high values to actions due to occasional factors, while advantage offers a more robust action assessment by taking the expectation of the future. 

\begin{algorithm}[t]
\caption{Advantage-Conditioned Transformer}
\label{algorithm}
\textbf{Input}: Initialized value networks $Q_\theta, V_\psi$, predictor network $c_\phi$, transformer \texttt{ACT}, offline dataset $\mathcal{D}$ 
\begin{algorithmic}[1]
\STATE // Training Phase
\FOR{$k=1, 2, \ldots, K$}
\STATE Update $Q_\theta$ and $V_\psi$ by Equation~\eqref{eq:dp} on dataset $\mathcal{D}$
\ENDFOR
\STATE Label the dataset $\mathcal{D}$ with estimated advantages to get the advantage-augmented dataset $\mathcal{D}^{\mathrm{A}}$
\STATE Learn \texttt{ACT} by minimizing Equation~\eqref{eq:act_objective} via stochastic gradient descent on dataset $\mathcal{D}^{\mathrm{A}}$
\STATE Learn $c_\phi$ by minimizing Equation~\eqref{eq:condition_objective} via stochastic gradient descent on dataset $\mathcal{D}^{\mathrm{A}}$
\STATE
\STATE // Test Phase
\STATE Initialize history $h=\emptyset$
\WHILE{episode not ended at timestep $t$}
    \STATE Let $A_t=c_\phi(s_t)$
    \STATE Execute $a_t=\texttt{ACT}(h, s_t, A_t)$
    \STATE Update $h=h\cup\{s_t, a_t\}$
\ENDWHILE
\end{algorithmic}
\end{algorithm}

\section{Related Work}
\textbf{Dealing with stochastic environments for DT. }
Numerous related works are dedicated to addressing the vulnerability of DT when confronted with stochastic environments.
ESPER~\cite{esper} uses adversarial clustering to learn trajectory representations disentangled from environmental stochasticity.
DoC~\cite{doc} attributes such failure to the fact that generative models make no distinction between the parts it can control~(agent actions) and those it cannot~(environmental transition).
Thus, DoC proposes to extract predictive representations for trajectories, while minimizing the mutual information between the representation and the environment transition.
A similar idea is also explored in~\citet{optimism_bias}, where the authors explicitly model the policy and the world model with two separate transformers.
\citet{rcsl_dp_analysis} provided a theoretical analysis of the return-conditioning regime, where they found the near-determinism of the environment is one of the conditions that return-conditioning can find the optimal policy. 

\textbf{Combining DP and DT. }Most of the offline RL methods employ dynamic programming to optimize policies, such as CQL~\cite{cql}, IQL~\cite{iql}, and TD3+BC~\cite{td3bc}.
As discussed in this paper as well as in \citet{rcsl_dp_analysis}, DP and DT bear distinct characteristics, for example, DP has no preference for the determinism of the environment, while DT may be more applicable in long-horizon tasks.
Several works have tried to combine DP and DT.
Among them, QDT~\cite{qdt} pre-computes conservative Q-values and V-values using CQL, and labels the offline data with the maximum of RTG and the conservative V-value. Our method differs from QDT in that, we compute the value estimations via an in-sample value iteration, which prevents instability and inaccuracy caused by bootstrapping from out-of-dataset actions. Moreover, the flexible advantage estimation also enables ACT to combine the best of both worlds.
Recently, EDT~\cite{edt} was proposed to vary the context length of DT during testing. When provided a shorter context, EDT can recover from bad history and switch to a higher rewarding action; provided a long context, EDT can stably recall the subsequent decisions.
In this way, EDT interpolates between trajectory stitching and behavior cloning from a new perspective. This work is complementary to ACT, and they can be seamlessly integrated to achieve better performance.

\section{Experimental Evaluations}\label{sec6}
Our assessments are designed to comprehensively analyze the efficacy of ACT when presented with diverse benchmarks and tasks encompassing a spectrum of characteristics and challenges, including deterministic, stochastic, and delayed reward tasks. We also conduct an ablation study on the use of network $c_\phi$ and the choice of transformer architecture.

\textbf{Deterministic Gym MuJoCo tasks. }We investigate the performance of ACT in the most widely studied Gym MuJoCo tasks. We focus on three domains, namely \textit{halfcheetah (hc)}, \textit{hopper (hp)}, and \textit{walker2d (wk)}, which are deterministic both in state transition and reward functions. We leverage the \textit{v2} datasets provided by D4RL~\cite{d4rl}, which includes three levels of quality: \textit{medium (med)}, \textit{medium-replay (med-rep)}, and \textit{medium-expert (med-exp)}. 

For this benchmark, we choose GAE(0) as the advantage estimator across all tasks to maximize the utility of dynamic programming. For $\sigma_1$, we sweep its value through $[0.5, 0.7]$. In order to mitigate the instability of TD learning~\cite{regulation}, we conduct model selection by choosing a value model that has the lowest training error on the offline dataset. See Appendix~B.2 for details. 

We contrast the performance of ACT against two groups of algorithms. The first group, including CQL~\cite{cql}, IQL~\cite{iql}, Onestep-RL~\cite{onestep_rl}, 10\%-BC, and RvS~\cite{rvs}, optimizes a Markovian policy that depends only on states. The second group, including DT~\cite{dt}, QDT~\cite{qdt}, and EDT~\cite{edt}, models the RL trajectories with sequence-modeling methods and learns a non-Markovian policy that depends on history. 

\begin{table*}[htbp]
\centering
\begin{small}
\scalebox{1.0}{
\begin{tabular}{ccrrrrr|rrrr}
\toprule[1.5pt]
\multicolumn{2}{c}{\multirow{2}{*}{Dataset}} & \multicolumn{5}{c}{Markovian}& \multicolumn{4}{c}{Non-Markovian} \\
\cmidrule(lr){3-7}\cmidrule(lr){8-11}

& &\multicolumn{1}{r}{CQL}   & \multicolumn{1}{r}{Onestep RL} & \multicolumn{1}{r}{IQL} &\multicolumn{1}{c}{10\% BC} & \multicolumn{1}{r}{RvS} &\multicolumn{1}{r}{DT}&\multicolumn{1}{r}{QDT}&\multicolumn{1}{r}{EDT}&  \multicolumn{1}{r}{ACT}\\ 
\midrule

\multicolumn{1}{c}{hc} & \multicolumn{1}{c}{\multirow{3}{*}{med}} &$44.0$\,\,& $48.4$\,\,   &$47.4$  &$42.5$  &$41.6$  & $42.6$   &$42.4\pm\,\,\, 0.5$ &$42.5\pm\,\, 0.9$& \colorbox{gray!30}{$\bm{49.1\pm\, 0.2}$} \\

\multicolumn{1}{c}{hp} &&$58.5$\,\, &$59.6$\,\,  & $66.2$   &$56.9$  &$60.2$  & $62.3$   &$60.7\pm\,\,\, 5.0$ &$63.5\pm\,\, 5.8$& \colorbox{gray!30}{$\bm{67.8\pm\, 5.5}$} \\

\multicolumn{1}{c}{wk} &&$72.5$\,\, & \colorbox{gray!30}{$81.8$}   &$78.3$  & $75.0$   &$71.7$  &$74.3$  &$63.7\pm\,\,\, 6.4$ &$72.8\pm\,\, 6.2$& $\bm{80.9\pm\, 0.4}$\,\,\\
\midrule

\multicolumn{1}{c}{hc} & \multicolumn{1}{c}{\multirow{3}{*}{med-rep}} & \colorbox{gray!30}{$45.5$} &$38.1$\,\,  &$44.2$  & $40.6$   &$38.0$  &$36.9$  &$32.8\pm\,\,\, 7.3$ &$37.8\pm\,\, 1.5$& $\bm{43.0\pm\, 0.4}$\,\,\\

\multicolumn{1}{c}{hp} &&$60.9$\,\, & $97.5$\,\,   &$94.7$  &$75.9$  &$73.5$  & $75.8$   &$38.7\pm26.7$ &$89.0\pm\,\, 8.3$& \colorbox{gray!30}{$\bm{98.4\pm\, 2.4}$} \\

\multicolumn{1}{c}{wk} && \colorbox{gray!30}{$77.2$} &$49.5$\,\, &$73.8$  &$62.5$  &$60.6$  & $59.4$   &$29.6\pm15.5$ & $\bm{74.8\pm4.9}$ &$56.1\pm10.9$\,\,  \\
\midrule

\multicolumn{1}{c}{hc} & \multicolumn{1}{c}{\multirow{3}{*}{med-exp}} &$91.6$\,\,& $93.4$\,\,   &$86.7$  & $92.9$   &$92.2$  &$86.3$  &$-$ &$-$ &\colorbox{gray!30}{$\bm{96.1\pm\, 1.4}$}\\

\multicolumn{1}{c}{hp} && $105.4$\,\, &$103.3$\,\, &$91.5$ & $110.9$  &$101.7$ &$104.2$ &$-$ &$-$& \colorbox{gray!30}{$\bm{111.5\pm\, 1.6}$} \\

\multicolumn{1}{c}{wk} &&$108.8$\,\,& $113.0$\,\,  &$109.6$ & $109.0$  &$106.0$ &$107.7$ &$-$ & $-$&\colorbox{gray!30}{$\bm{113.3\pm \, 0.4}$}\\

\bottomrule[1.5pt]
\end{tabular}
}
\end{small}
\caption{Normalized score on deterministic Gym MuJoCo tasks, with datasets from D4RL. The performances of QDT and EDT are taken from their original papers, and the numbers for other baselines are taken from ~\cite{xql}. For ACT and DT, we use our own implementations and report the average performance and the standard deviation of the final checkpoint across 10 evaluation episodes and 5 seeds. We bold the highest score among sequence-modeling methods, and add background shading to the highest score among all methods.}
\label{tab:deterministic_gym_perf}
\end{table*}

\begin{figure}[htbp]
    \centering
    \includegraphics[width=1.0\linewidth]{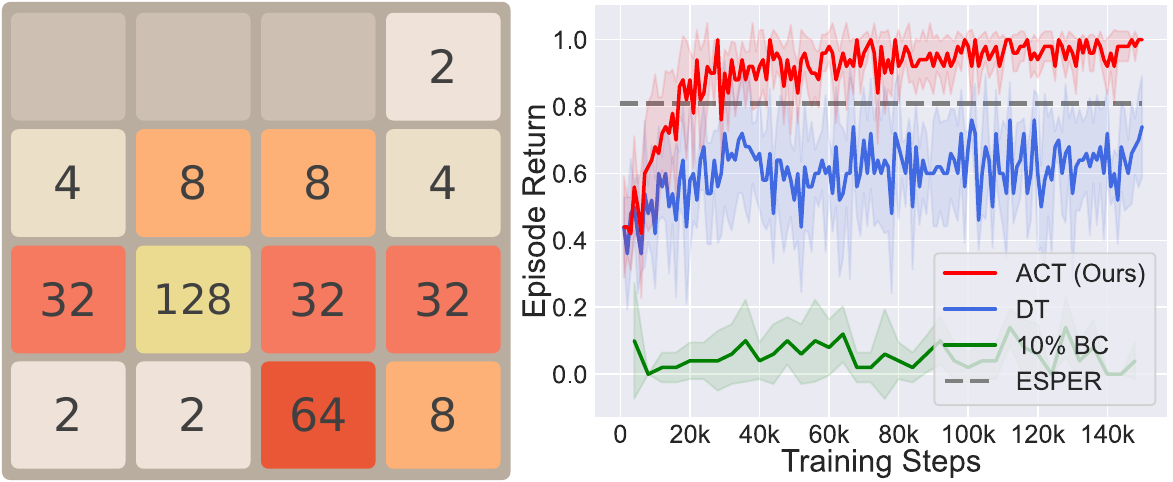}
    \caption{Performance on the 2048 game. We report the average and std of the performance across 5 independent runs and mark ESPER's score as the grey dotted line. }
    \label{fig:2048_perf}
\end{figure}

The results are listed in Table~\ref{tab:deterministic_gym_perf}. We find that ACT achieves the highest score in 8 out of 9 tasks compared to other methods that learn non-Markovian policies, including QDT and EDT which share a similar motivation with us, i.e., improving DT with the ability of trajectory stitching. The improvement over these methods underscores the efficacy of advantage conditioning. If we extend the comparison to all methods, ACT still manages to outperform in 6 out of 9 tasks. This demonstrates that ACT not only exploits the power of DP but also gains benefits from the sequence modeling technique, thus combining the best of both worlds.

\textbf{Stochastic benchmarks. }Another important property of ACT is its robustness against the stochasticity of the environments. To validate this, we choose to assess ACT, DT, QDT, ESPER~\cite{esper}, and 10\%-BC on stochastic benchmarks. Among them, ESPER is an algorithm that enhances DT to cope with environmental stochasticity. In the following evaluations, we select IAE as the advantage estimator, and $\sigma_1$ is kept to 0.5 by default. 

As a sanity check, we first evaluate the algorithms on the 2048 game. This environment is stochastic in that, after each action which moves all the tiles along some direction and merges them if their numbers are equal, a random 2 or 4 will be placed in a random empty grid. The agent is given a reward of 1 once it manages to produce a tile of 128, and the episode will be ended. Thus, the maximum return in this environment is 1. Further introduction about this environment and the dataset can be found in Appendix~B.3. 
Figure~\ref{fig:2048_perf} depicts the performance curve as the training proceeds. While DT and 10\%-BC exploit the occasional success in the dataset and converge to suboptimal policies, ACT and ESPER show robustness to stochasticity. Besides, ACT outperforms ESPER and converges to 100\% success rate to produce 128, validating the power of advantage conditioning. 

We also created a more sophisticated benchmark, by reusing the Gym MuJoCo tasks and following \citet{doc} to add noise to the agent's action before passing it to the simulator. More details can be found in Appendix~B.4. The results are illustrated in Figure~\ref{fig:stoc_mujoco_perf}. As expected, we observe that 10\%-BC severely overfits the high-return trajectories in the dataset, albeit with a sharp drop in test scores as the training proceeds. DT and QDT yield a comparatively robust policy, while their performances are still inferior to ACT. This further justifies the validity of ACT in stochastic control tasks. 

\begin{table}[htbp]
    \centering
    \scalebox{1.0}{
    \begin{small}
    \begin{tabular}{c|p{1.0cm}p{1.0cm}p{1.2cm}p{1.2cm}}
    \toprule[1.0pt]
        Dataset & CQL & IQL & DT & ACT\\
        \midrule
        hp-med & $ 46.1_{\pm2.9} $ & $ 32.4_{\pm7.7} $ & $ \bm{65.7_{\pm1.4}} $ & $ 49.4_{\pm1.5} $\\
        hp-med-exp & $ 0.8_{\pm0.3} $ & $ 97.6_{\pm14.8} $ & $ \bm{106.6_{\pm3.7}} $ & $ 95.4_{\pm13.9} $\\
        wk-med & $ -0.3_{\pm0.1} $ & $ 53.2_{\pm6.7} $ & $ 72.1_{\pm4.3} $ & $ \bm{73.8_{\pm1.9}} $\\
        wk-med-exp & $ 7.0_{\pm6.4} $ & $ 67.4_{\pm22.1} $ & $ 107.4_{\pm0.3} $ & $ \bm{108.1_{\pm0.2}} $\\
        \midrule
        Average & $13.4$&$62.7$&$\bm{88.0}$& $81.7$\\
        \bottomrule[1.0pt]
    \end{tabular}
    \end{small}
    }
    \caption{Normalized score on delayed reward tasks. The average and std are taken across 4 independent runs.}
    \label{tab:delayed_perf}
\end{table}

\begin{figure*}[htbp]
    \centering
    \includegraphics[width=1.0\linewidth]{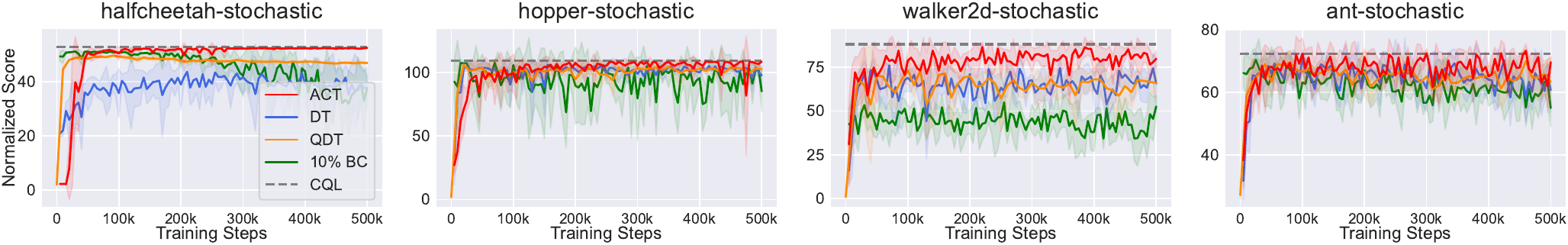}
    \caption{Performance curve on the stochastic Gym MuJoCo tasks as the training proceeds. We report the average and the standard deviation of the performance across 4 independent runs. }
    \label{fig:stoc_mujoco_perf}
\end{figure*}

\begin{figure}[htbp]
    \centering
    \includegraphics[width=0.98\linewidth]{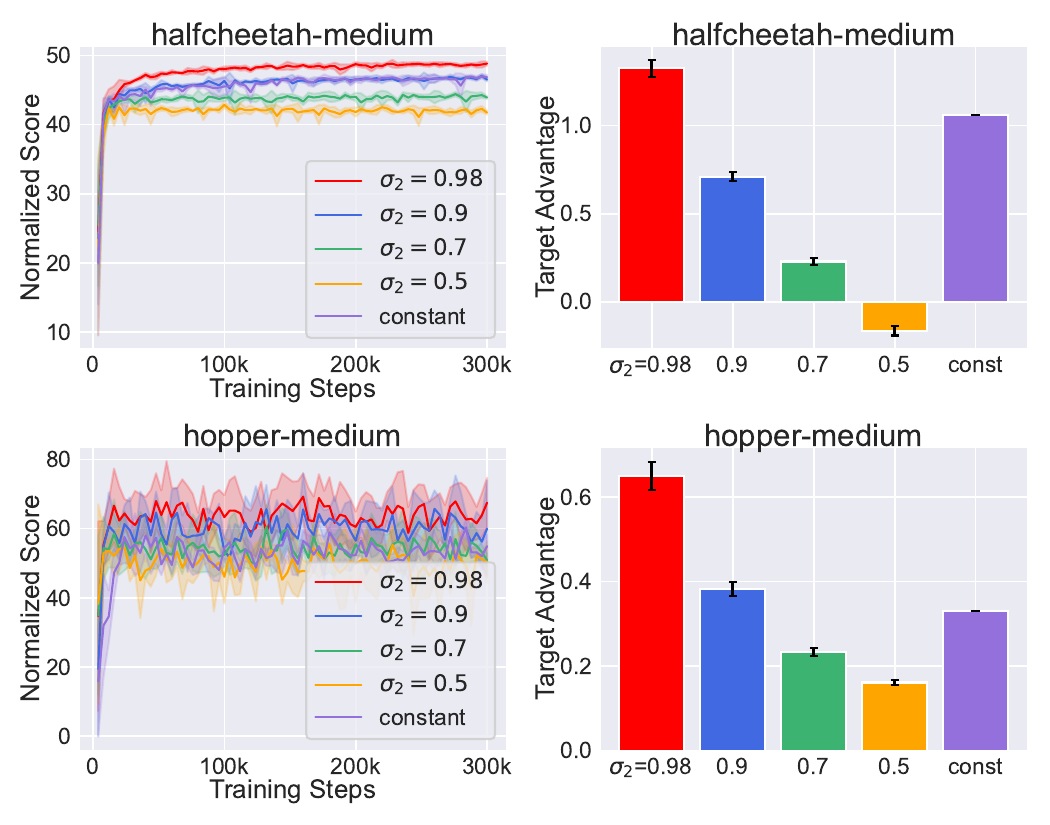}
    \caption{Ablation study on the effect of $\sigma_2$, using datasets from D4RL. The left column depicts the performance of each variant as the training proceeds, and the right column depicts the target advantages given by $c_\phi$. The results are taken from 4 independent runs.}
    \label{fig:abl_sigma2}
\end{figure}

\textbf{Delayed reward tasks. }As unveiled in previous literature~\cite{dt, qdt}, DT has an advantage over conventional RL methods when the reward is sparse and thus long-term credit assignment is required. In the following part, we aim to investigate whether ACT still retains such ability. We once again revise the D4RL datasets, deferring the rewards for each trajectory until the final timestep. We exclude the \textit{hc} and \textit{med-rep} datasets as they contain time-out trajectories which are not applicable for delayed-reward settings. For the advantage estimator, we choose GAE(1) for its resemblance to the RTGs. The results are listed in Table~\ref{tab:delayed_perf}. While DP-based methods like CQL and IQL suffer from degeneration, DT upholds its performance and is minimally affected by reward sparsity. Although ACT is not entirely spared from the influence, it still preserves benefits from the sequence modeling architecture and outperforms CQL and IQL. 

\textbf{Ablation study. }
Finally, we investigate the effects of the design choices of ACT through ablation studies. Our first ablation focuses on the effect of $\sigma_2$, which determines the expectile to approximate by $c_\phi$. In the above experiments, we set the $\sigma_2$ to $0.98$, and in this ablation study, we additionally test with $0.5$, $0.7$, and $0.9$ to analyze the actual effect of this parameter. According to the discussions in Section~\ref{sec4.4:benefits}, providing ACT with a positive advantage already induces policy improvement. Therefore we also include a variant of ACT which uses a constant positive value as the target advantage for all states. The positive value is set as the average of the positive advantages in the offline dataset.
The results illustrated in Figure~\ref{fig:abl_sigma2} suggest a close relationship between the value of $\sigma_2$ and the final performance.
When $\sigma_2$ is set to higher values, the predictor network $c_\phi$ tends to give higher desired advantages $\hat{A}$ during test time, and ACT would retrieve better actions in response. 
Giving a positive constant value also brings about improvement in certain environments such as \textit{hc-med}, but in \textit{hp-med} it performs substantially worse than using a predictor network. Comparisons on more datasets are deferred to Appendix~C.2.  

\begin{figure}[htbp]
    \centering
    \includegraphics[width=1.0\linewidth]{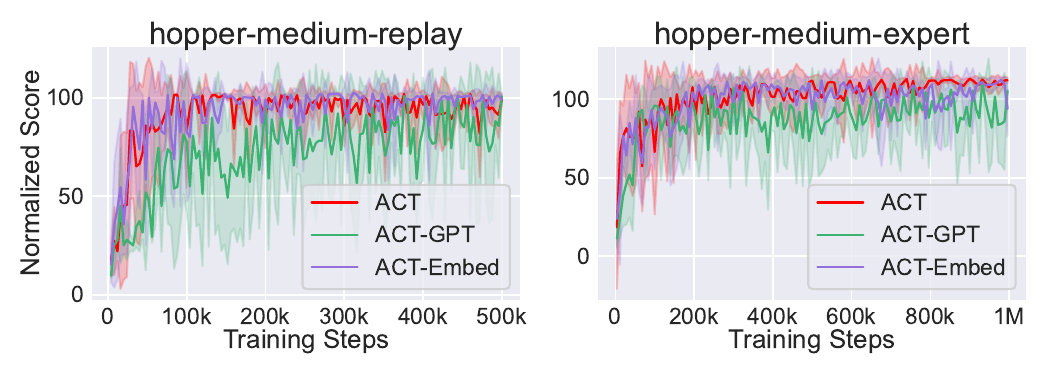}
    \caption{Ablation study on the transformer architecture, using datasets from D4RL with 4 independent runs.}
    \label{fig:abl_arch}
\end{figure}

We also carried out an analysis of the choice of transformer architecture. ACT differs from DT in two ways: 1) ACT employs an encoder-decoder structure while DT is a decoder-only transformer, and 2) ACT uses sinusoidal encoding rather than learnable embedding to prevent overfitting. Thus, we additionally implement two variants of ACT, one with GPT-2 architecture~(\texttt{ACT-GPT}) and another using learnable positional embedding~(\texttt{ACT-Embed}), and keep other hyper-parameters the same with ACT. In Figure~\ref{fig:abl_arch}, we select two datasets of the hopper task for comparison and observe that \texttt{ACT-GPT} has undergone significant oscillation and degradation in its performance, implying that the direct application of GPT-2 structure to our algorithm is not effective. The other variant, \texttt{ACT-Embed}, suffers from instabilities in tasks like \textit{hp-med-exp} in the late training, although the overall gap is minor. 

\section{Conclusion and Future Work}
In this paper, we introduce ACT, which enhances popular sequence modeling techniques with dynamic programming to address their limitations in offline policy optimization. We achieve this by conditioning the transformer architecture on the estimated advantages. 
By selectively choosing the type of advantage estimator, our framework can be applied to a variety of tasks, demonstrating consistent performance improvements and generalizability. Our framework also leaves room for the incorporation of self-supervised learning techniques and has the potential to be extended to multi-task settings and to a larger training scale. We will continue this line of research and investigate these topics in the future. 

\section*{Acknowledgements}
This work is supported by the National Science Foundation of China (62276126, 62250069), the Natural Science Foundation of Jiangsu (BK20221442), and the Fundamental Research Funds for the Central Universities (14380010).

\bibliography{aaai24}

\begin{thebibliography}{53}
\providecommand{\natexlab}[1]{#1}

\bibitem[{An et~al.(2021)An, Moon, Kim, and Song}]{edac}
An, G.; Moon, S.; Kim, J.; and Song, H.~O. 2021.
\newblock Uncertainty-Based Offline Reinforcement Learning with Diversified
  Q-Ensemble.
\newblock In \emph{Advances in Neural Information Processing Systems
  (NeurIPS)}, 7436--7447.

\bibitem[{Bai et~al.(2022)Bai, Wang, Yang, Deng, Garg, Liu, and Wang}]{pbrl}
Bai, C.; Wang, L.; Yang, Z.; Deng, Z.; Garg, A.; Liu, P.; and Wang, Z. 2022.
\newblock Pessimistic Bootstrapping for Uncertainty-Driven Offline
  Reinforcement Learning.
\newblock In \emph{International Conference on Learning Representations
  (ICLR)}.

\bibitem[{Brandfonbrener et~al.(2022)Brandfonbrener, Bietti, Buckman, Laroche,
  and Bruna}]{rcsl_dp_analysis}
Brandfonbrener, D.; Bietti, A.; Buckman, J.; Laroche, R.; and Bruna, J. 2022.
\newblock When Does Return-conditioned Supervised Learning Work for Offline
  Reinforcement Learning?
\newblock In \emph{Advances in Neural Information Processing Systems
  (NeurIPS)}, 1542--1553.

\bibitem[{Brandfonbrener et~al.(2021)Brandfonbrener, Whitney, Ranganath, and
  Bruna}]{onestep_rl}
Brandfonbrener, D.; Whitney, W.; Ranganath, R.; and Bruna, J. 2021.
\newblock Offline {RL} Without Off-Policy Evaluation.
\newblock In \emph{Advances in Neural Information Processing Systems
  (NeurIPS)}, 4933--4946.

\bibitem[{Brown et~al.(2020)Brown, Mann, Ryder, Subbiah, Kaplan, Dhariwal,
  Neelakantan, Shyam, Sastry, Askell, Agarwal, Herbert{-}Voss, Krueger,
  Henighan, Child, Ramesh, Ziegler, Wu, Winter, Hesse, Chen, Sigler, Litwin,
  Gray, Chess, Clark, Berner, McCandlish, Radford, Sutskever, and
  Amodei}]{gpt3}
Brown, T.~B.; Mann, B.; Ryder, N.; Subbiah, M.; Kaplan, J.; Dhariwal, P.;
  Neelakantan, A.; Shyam, P.; Sastry, G.; Askell, A.; Agarwal, S.;
  Herbert{-}Voss, A.; Krueger, G.; Henighan, T.; Child, R.; Ramesh, A.;
  Ziegler, D.~M.; Wu, J.; Winter, C.; Hesse, C.; Chen, M.; Sigler, E.; Litwin,
  M.; Gray, S.; Chess, B.; Clark, J.; Berner, C.; McCandlish, S.; Radford, A.;
  Sutskever, I.; and Amodei, D. 2020.
\newblock Language Models are Few-shot Learners.
\newblock In \emph{Advances in Neural Information Processing Systems
  (NeurIPS)}.

\bibitem[{Carroll et~al.(2022)Carroll, Paradise, Lin, Georgescu, Sun, Bignell,
  Milani, Hofmann, Hausknecht, Dragan, and Devlin}]{unimask}
Carroll, M.; Paradise, O.; Lin, J.; Georgescu, R.; Sun, M.; Bignell, D.;
  Milani, S.; Hofmann, K.; Hausknecht, M.~J.; Dragan, A.~D.; and Devlin, S.
  2022.
\newblock Uni[MASK]: Unified Inference in Sequential Decision Problems.
\newblock In \emph{Advances in Neural Information Processing Systems
  (NeurIPS)}.

\bibitem[{Chen et~al.(2021)Chen, Lu, Rajeswaran, Lee, Grover, Laskin, Abbeel,
  Srinivas, and Mordatch}]{dt}
Chen, L.; Lu, K.; Rajeswaran, A.; Lee, K.; Grover, A.; Laskin, M.; Abbeel, P.;
  Srinivas, A.; and Mordatch, I. 2021.
\newblock Decision Transformer: Reinforcement Learning via Sequence Modeling.
\newblock In \emph{Advances in Neural Information Processing Systems
  (NeurIPS)}, 15084--15097.

\bibitem[{Emmons et~al.(2022)Emmons, Eysenbach, Kostrikov, and Levine}]{rvs}
Emmons, S.; Eysenbach, B.; Kostrikov, I.; and Levine, S. 2022.
\newblock RvS: What is Essential for Offline {RL} via Supervised Learning?
\newblock In \emph{International Conference on Learning Representations
  (ICLR)}.

\bibitem[{Fu et~al.(2020)Fu, Kumar, Nachum, Tucker, and Levine}]{d4rl}
Fu, J.; Kumar, A.; Nachum, O.; Tucker, G.; and Levine, S. 2020.
\newblock {D4RL:} Datasets for Deep Data-Driven Reinforcement Learning.
\newblock \emph{arXiv preprint arXiv:2004.07219}.

\bibitem[{Fujimoto and Gu(2021)}]{td3bc}
Fujimoto, S.; and Gu, S.~S. 2021.
\newblock A Minimalist Approach to Offline Reinforcement Learning.
\newblock In \emph{Advances in Neural Information Processing Systems
  (NeurIPS)}, 20132--20145.

\bibitem[{Fujimoto, Meger, and Precup(2019)}]{bcq}
Fujimoto, S.; Meger, D.; and Precup, D. 2019.
\newblock Off-Policy Deep Reinforcement Learning without Exploration.
\newblock In \emph{International Conference on Machine Learning (ICML)},
  2052--2062.

\bibitem[{Fujimoto, van Hoof, and Meger(2018)}]{td3}
Fujimoto, S.; van Hoof, H.; and Meger, D. 2018.
\newblock Addressing Function Approximation Error in Actor-Critic Methods.
\newblock In \emph{International Conference on Machine Learning (ICML)},
  1582--1591.

\bibitem[{Garg et~al.(2023)Garg, Hejna, Geist, and Ermon}]{xql}
Garg, D.; Hejna, J.; Geist, M.; and Ermon, S. 2023.
\newblock Extreme Q-Learning: MaxEnt {RL} without Entropy.
\newblock In \emph{International Conference on Learning Representations
  (ICLR)}.

\bibitem[{Haarnoja et~al.(2018)Haarnoja, Zhou, Abbeel, and Levine}]{sac}
Haarnoja, T.; Zhou, A.; Abbeel, P.; and Levine, S. 2018.
\newblock Soft Actor-Critic: Off-Policy Maximum Entropy Deep Reinforcement
  Learning with a Stochastic Actor.
\newblock In \emph{International Conference on Machine Learning (ICML)},
  1856--1865.

\bibitem[{Hejna, Gao, and Sadigh(2023)}]{dwsl}
Hejna, J.; Gao, J.; and Sadigh, D. 2023.
\newblock Distance Weighted Supervised Learning for Offline Interaction Data.
\newblock \emph{arXiv preprint arXiv:2304.13774}.

\bibitem[{Kakade and Langford(2002)}]{policy_difference_lemma}
Kakade, S.; and Langford, J. 2002.
\newblock Approximately Optimal Approximate Reinforcement Learning.
\newblock In \emph{International Conference on Machine Learning (ICML)},
  267--274.

\bibitem[{Kostrikov, Nair, and Levine(2022)}]{iql}
Kostrikov, I.; Nair, A.; and Levine, S. 2022.
\newblock Offline Reinforcement Learning with Implicit Q-Learning.
\newblock In \emph{International Conference on Learning Representations
  (ICLR)}.

\bibitem[{Kumar et~al.(2019)Kumar, Fu, Soh, Tucker, and Levine}]{bear}
Kumar, A.; Fu, J.; Soh, M.; Tucker, G.; and Levine, S. 2019.
\newblock Stabilizing Off-Policy Q-Learning via Bootstrapping Error Reduction.
\newblock In \emph{Advances in Neural Information Processing Systems
  (NearIPS)}, 11761--11771.

\bibitem[{Kumar et~al.(2021{\natexlab{a}})Kumar, Hong, Singh, and
  Levine}]{why_offlinerl_is_better}
Kumar, A.; Hong, J.; Singh, A.; and Levine, S. 2021{\natexlab{a}}.
\newblock Should I Run Offline Reinforcement Learning or Behavioral Cloning?
\newblock In \emph{International Conference on Learning Representations
  (ICLR)}.

\bibitem[{Kumar, Peng, and Levine(2019)}]{rcp}
Kumar, A.; Peng, X.~B.; and Levine, S. 2019.
\newblock Reward-Conditioned Policies.
\newblock \emph{arXiv preprint arXiv:1912.13465}.

\bibitem[{Kumar et~al.(2021{\natexlab{b}})Kumar, Singh, Tian, Finn, and
  Levine}]{robotics1}
Kumar, A.; Singh, A.; Tian, S.; Finn, C.; and Levine, S. 2021{\natexlab{b}}.
\newblock A Workflow for Offline Model-Free Robotic Reinforcement Learning.
\newblock In \emph{Conference on Robot Learning (CORL)}, 417--428.

\bibitem[{Kumar et~al.(2020)Kumar, Zhou, Tucker, and Levine}]{cql}
Kumar, A.; Zhou, A.; Tucker, G.; and Levine, S. 2020.
\newblock Conservative Q-Learning for Offline Reinforcement Learning.
\newblock In \emph{Advances in Neural Information Processing Systems
  (NeurIPS)}, 1179--1191.

\bibitem[{Levine et~al.(2020)Levine, Kumar, Tucker, and Fu}]{levine_survey}
Levine, S.; Kumar, A.; Tucker, G.; and Fu, J. 2020.
\newblock Offline Reinforcement Learning: Tutorial, Review, and Perspectives on
  Open Problems.
\newblock \emph{arXiv preprint arXiv:2005.01643}.

\bibitem[{Li et~al.(2023)Li, Kumar, Kostrikov, and Levine}]{regulation}
Li, Q.; Kumar, A.; Kostrikov, I.; and Levine, S. 2023.
\newblock Efficient Deep Reinforcement Learning Requires Regulating
  Overfitting.
\newblock In \emph{International Conference on Learning Representations
  (ICLR)}.

\bibitem[{Liu et~al.(2018)Liu, Zhai, Zhang, Zhong, Zhou, Zhang, and
  Xu}]{liu2018}
Liu, Q.; Zhai, J.; Zhang, Z.; Zhong, S.; Zhou, Q.; Zhang, P.; and Xu, J. 2018.
\newblock A Survey on Deep Reinforcement Learning.
\newblock \emph{Chinese Journal of Computers}, 41(1): 1--27.

\bibitem[{Paster, McIlraith, and Ba(2022)}]{esper}
Paster, K.; McIlraith, S.; and Ba, J. 2022.
\newblock You Can’t Count on Luck: Why Decision Transformers and RvS Fail in
  Stochastic Environments.
\newblock In \emph{Advances in Neural Information Processing Systems
  (NeurIPS)}, 38966--38979.

\bibitem[{Qin et~al.(2022)Qin, Zhang, Gao, Chen, Li, Zhang, and Yu}]{neorl}
Qin, R.-J.; Zhang, X.; Gao, S.; Chen, X.-H.; Li, Z.; Zhang, W.; and Yu, Y.
  2022.
\newblock NeoRL: A near real-world benchmark for offline reinforcement
  learning.
\newblock \emph{Advances in Neural Information Processing Systems (NeurIPS)},
  24753--24765.

\bibitem[{Radford et~al.(2019)Radford, Wu, Child, Luan, Amodei, and
  Sutskever}]{gpt2}
Radford, A.; Wu, J.; Child, R.; Luan, D.; Amodei, D.; and Sutskever, I. 2019.
\newblock Language Models are Unsupervised Multitask Learners.
\newblock In \emph{OpenAI Blog}.

\bibitem[{Ran et~al.(2023)Ran, Li, Zhang, Zhang, and Yu}]{prdc}
Ran, Y.; Li, Y.; Zhang, F.; Zhang, Z.; and Yu, Y. 2023.
\newblock Policy Regularization with Dataset Constraint for Offline
  Reinforcement Learning.
\newblock In \emph{International Conference on Machine Learning (ICML)},
  28701--28717.

\bibitem[{Schmidhuber(2019)}]{udrl}
Schmidhuber, J. 2019.
\newblock Reinforcement Learning Upside Down: Don't Predict Rewards - Just Map
  Them to Actions.
\newblock \emph{arXiv preprint arXiv:1912.02875}.

\bibitem[{Schulman et~al.(2015)Schulman, Levine, Abbeel, Jordan, and
  Moritz}]{trpo}
Schulman, J.; Levine, S.; Abbeel, P.; Jordan, M.; and Moritz, P. 2015.
\newblock Trust region policy optimization.
\newblock In \emph{International Conference on Machine Learning (ICML)},
  1889--1897.

\bibitem[{Schulman et~al.(2016)Schulman, Moritz, Levine, Jordan, and
  Abbeel}]{gae}
Schulman, J.; Moritz, P.; Levine, S.; Jordan, M.~I.; and Abbeel, P. 2016.
\newblock High-Dimensional Continuous Control Using Generalized Advantage
  Estimation.
\newblock In \emph{International Conference on Learning Representations
  (ICLR)}.

\bibitem[{Schulman et~al.(2017)Schulman, Wolski, Dhariwal, Radford, and
  Klimov}]{ppo}
Schulman, J.; Wolski, F.; Dhariwal, P.; Radford, A.; and Klimov, O. 2017.
\newblock Proximal Policy Optimization Algorithms.
\newblock \emph{arXiv preprint arXiv:1707.06347}.

\bibitem[{Shiranthika et~al.(2022)Shiranthika, Chen, Wang, Yang, Sudantha, and
  Li}]{healthcare1}
Shiranthika, C.; Chen, K.; Wang, C.; Yang, C.; Sudantha, B.~H.; and Li, W.
  2022.
\newblock Supervised Optimal Chemotherapy Regimen Based on Offline
  Reinforcement Learning.
\newblock \emph{{IEEE} Journal of Biomedical and Health Informatics}, 26(9):
  4763--4772.

\bibitem[{Sun et~al.(2023{\natexlab{a}})Sun, Ma, Madaan, Bonatti, Huang, and
  Kapoor}]{smart}
Sun, Y.; Ma, S.; Madaan, R.; Bonatti, R.; Huang, F.; and Kapoor, A.
  2023{\natexlab{a}}.
\newblock {SMART:} Self-supervised Multi-task pretrAining with contRol
  Transformers.
\newblock In \emph{International Conference on Learning Representations
  (ICLR)}.

\bibitem[{Sun et~al.(2023{\natexlab{b}})Sun, Zhang, Jia, Lin, Ye, and
  Yu}]{mobile}
Sun, Y.; Zhang, J.; Jia, C.; Lin, H.; Ye, J.; and Yu, Y. 2023{\natexlab{b}}.
\newblock Model-Bellman Inconsistency for Model-based Offline Reinforcement
  Learning.
\newblock In \emph{International Conference on Machine Learning (ICML)},
  33177--33194.

\bibitem[{Sutton et~al.(1999)Sutton, McAllester, Singh, and
  Mansour}]{policy_gradient}
Sutton, R.~S.; McAllester, D.~A.; Singh, S.; and Mansour, Y. 1999.
\newblock Policy Gradient Methods for Reinforcement Learning with Function
  Approximation.
\newblock In \emph{Advances in Neural Information Processing Systems (NIPS)},
  1057--1063.

\bibitem[{Tarasov et~al.(2022)Tarasov, Nikulin, Akimov, Kurenkov, and
  Kolesnikov}]{corl}
Tarasov, D.; Nikulin, A.; Akimov, D.; Kurenkov, V.; and Kolesnikov, S. 2022.
\newblock {CORL}: Research-oriented Deep Offline Reinforcement Learning
  Library.
\newblock In \emph{Offline RL Workshop: Offline RL as a ``Launchpad''}.

\bibitem[{Vaswani et~al.(2017)Vaswani, Shazeer, Parmar, Uszkoreit, Jones,
  Gomez, Kaiser, and Polosukhin}]{attention_is_all_you_need}
Vaswani, A.; Shazeer, N.; Parmar, N.; Uszkoreit, J.; Jones, L.; Gomez, A.~N.;
  Kaiser, L.; and Polosukhin, I. 2017.
\newblock Attention is All you Need.
\newblock In \emph{Advances in Neural Information Processing Systems
  (NeurIPS)}, 5998--6008.

\bibitem[{Villaflor et~al.(2022)Villaflor, Huang, Pande, Dolan, and
  Schneider}]{optimism_bias}
Villaflor, A.~R.; Huang, Z.; Pande, S.; Dolan, J.~M.; and Schneider, J. 2022.
\newblock Addressing Optimism Bias in Sequence Modeling for Reinforcement
  Learning.
\newblock In \emph{International Conference on Machine Learning (ICML)},
  22270--22283.

\bibitem[{Wen et~al.(2023)Wen, Lin, Wang, Yang, Wen, Mai, Wang, Zhang, and
  Zhang}]{wen2023}
Wen, M.; Lin, R.; Wang, H.; Yang, Y.; Wen, Y.; Mai, L.; Wang, J.; Zhang, H.;
  and Zhang, W. 2023.
\newblock Large Sequence Models for Sequential Decision-Making: A Survey.
\newblock \emph{Frontiers of Computer Science}, 17(6): Article Number 176349.

\bibitem[{Wu and Zhang(2023)}]{wu2023}
Wu, C.; and Zhang, Z. 2023.
\newblock Surfing Information: The Challenge of Intelligent Decision-Making.
\newblock \emph{Intelligent Computing}, 2: Article 0041.

\bibitem[{Wu et~al.(2023)Wu, Majumdar, Stone, Lin, Mordatch, Abbeel, and
  Rajeswaran}]{mtm}
Wu, P.; Majumdar, A.; Stone, K.; Lin, Y.; Mordatch, I.; Abbeel, P.; and
  Rajeswaran, A. 2023.
\newblock Masked Trajectory Models for Prediction, Representation, and Control.
\newblock In \emph{International Conference on Machine Learning (ICML)},
  37607--37623.

\bibitem[{Wu, Wang, and Hamaya(2023)}]{edt}
Wu, Y.; Wang, X.; and Hamaya, M. 2023.
\newblock Elastic Decision Transformer.
\newblock \emph{arXiv preprint arXiv:2307.02484}.

\bibitem[{Xiao et~al.(2023)Xiao, Wang, Pan, White, and White}]{inac}
Xiao, C.; Wang, H.; Pan, Y.; White, A.; and White, M. 2023.
\newblock The In-Sample Softmax for Offline Reinforcement Learning.
\newblock In \emph{International Conference on Learning Representations
  (ICLR)}.

\bibitem[{Yamagata, Khalil, and Santos-Rodriguez(2023)}]{qdt}
Yamagata, T.; Khalil, A.; and Santos-Rodriguez, R. 2023.
\newblock Q-learning Decision Transformer: Leveraging Dynamic Programming for
  Conditional Sequence Modelling in Offline RL.
\newblock In \emph{International Conference on Machine Learning (ICML)},
  38989--39007.

\bibitem[{Yang et~al.(2023)Yang, Schuurmans, Abbeel, and Nachum}]{doc}
Yang, S.; Schuurmans, D.; Abbeel, P.; and Nachum, O. 2023.
\newblock Dichotomy of Control: Separating What You Can Control from What You
  Cannot.
\newblock In \emph{International Conference on Learning Representations
  (ICLR)}.

\bibitem[{Yu et~al.(2021)Yu, Kumar, Rafailov, Rajeswaran, Levine, and
  Finn}]{combo}
Yu, T.; Kumar, A.; Rafailov, R.; Rajeswaran, A.; Levine, S.; and Finn, C. 2021.
\newblock {COMBO:} Conservative Offline Model-Based Policy Optimization.
\newblock In \emph{Advances in Neural Information Processing Systems
  (NeurIPS)}, 28954--28967.

\bibitem[{Yu et~al.(2020)Yu, Thomas, Yu, Ermon, Zou, Levine, Finn, and
  Ma}]{mopo}
Yu, T.; Thomas, G.; Yu, L.; Ermon, S.; Zou, J.~Y.; Levine, S.; Finn, C.; and
  Ma, T. 2020.
\newblock {MOPO:} Model-based Offline Policy Optimization.
\newblock In \emph{Advances in Neural Information Processing Systems
  (NeurIPS)}.

\bibitem[{Zhang et~al.(2023)Zhang, Mao, Wang, He, Xu, and Ji}]{iac}
Zhang, H.; Mao, Y.; Wang, B.; He, S.; Xu, Y.; and Ji, X. 2023.
\newblock In-sample Actor Critic for Offline Reinforcement Learning.
\newblock In \emph{International Conference on Learning Representations
  (ICLR)}.

\bibitem[{Zheng, Zhang, and Grover(2022)}]{onlinedt}
Zheng, Q.; Zhang, A.; and Grover, A. 2022.
\newblock Online Decision Transformer.
\newblock In \emph{International Conference on Machine Learning (ICML)},
  27042--27059.

\bibitem[{Zhou et~al.(2024)Zhou, Gao, Zhang, and Yu}]{zhou2024}
Zhou, R.; Gao, C.; Zhang, Z.; and Yu, Y. 2024.
\newblock Generalizable Task Representation Learning for Offline
  Meta-Reinforcement Learning with Data Limitations.
\newblock In \emph{AAAI Conference on Artificial Intelligence (AAAI)}.

\bibitem[{Zhou, Zhang, and Yu(2023)}]{zhou2023}
Zhou, R.; Zhang, Z.; and Yu, Y. 2023.
\newblock Model-based Offline Weighted Policy Optimization.
\newblock In \emph{AAAI Conference on Artificial Intelligence (AAAI)},
  16392--16393.

\end{thebibliography}

\clearpage
\appendix
\section{Remarks on Test-time Policy Improvement}\label{app:test-time_policy_improvement}
In this section, we remark on how ACT and DT improve their policies with the offline dataset. The training phase of ACT and DT can be interpreted as memorizing the offline data together with its hindsight information, either RTG or the advantages. DT and ACT improve the policy by conditionally generating an above-average action during the testing phase, which sets them apart from traditional RL algorithms which optimize policies during training. Such test-time improvement is done by retrieving actions with the desired hindsight information. In DT, the retrieval is accomplished by providing DT with a slightly over-estimated RTG. In ACT, the relationship between advantages and policy performance enables us to make the action generation grounded on theory. 

Recall the Performance Difference Lemma:
\begin{lemma}{(Performance Difference Lemma~\cite{policy_difference_lemma})}
Consider an infinite horizon MDP $\mathcal{M}=\langle\mathcal{S}, \mathcal{A}, T, r, \rho_0, \gamma\rangle$. We have
$$
\eta(\pi)-\eta(\pi')=\mathbb{E}_{\tau\sim\pi}\left[\sum_{t=0}^\infty \gamma^t A^{\pi'}(s_t, a_t)\right]. 
$$
\end{lemma}

In ACT, we train the value functions via expectile regression, which smoothly interpolates between estimating $(Q^\beta, V^\beta)$ and $(Q_\beta^*, V_\beta^*)$. We take the former one as an example, which corresponds to setting $\sigma_1=0.5$. In such case, $\pi'=\beta$. Let $\epsilon=\max_{s,a}|\hat{A}^\beta(s, a) - A^\beta(s, a)|$ be the maximum error between the estimated advantage $\hat A^\beta$ and the ground truth $A^\beta$. Then, we have
$$
\begin{aligned}
&\eta(\pi)-\eta(\beta) \\
&= \mathbb{E}_{\tau\sim\pi}\left[\sum_{t=0}^\infty \gamma^t A^{\beta}(s_t, a_t)\right]\\
&=\mathbb{E}_{\tau\sim\pi}\left[\sum_{t=0}^\infty \gamma^t \hat A^{\beta}(s_t, a_t)\right]\\
&\quad\quad+\mathbb{E}_{\tau\sim\pi}\left[\sum_{t=0}^\infty \gamma^t \Big(A^{\beta}(s_t, a_t)-\hat A^{\beta}(s_t, a_t)\Big)\right]\\
&\geq\mathbb{E}_{\tau\sim\pi}\left[\sum_{t=0}^\infty \gamma^t \hat A^{\beta}(s_t, a_t)\right] -\frac{\epsilon}{1-\gamma}.\\
\end{aligned}
$$
We denote $\Delta_c$ as the resulting performance difference lower bound $\mathbb{E}_{\tau\sim\pi}[\sum_{t=0}^\infty \gamma^t \hat A^{\beta}(s_t, a_t)] -\frac{\epsilon}{1-\gamma}$.
An intuitive interpretation of the above mathematics is that, if during testing we consistently provide ACT with a positive target advantage, we will improve the performance over the behavior policy $\beta$ by a margin of $\Delta_c$. 

However, $\Delta_c$ is not guaranteed to be positive since it is also coupled with the error $\epsilon$. The advantage estimation can be very poor in the limit of $\sigma_1\to1$.
We do observe that when setting $\sigma_1=0.7$, the performance of ACT drops slightly in some tasks that we are concerned about. We speculate that the drop can be explained by the mismatch in the data distribution, since when $\sigma_1=0.7$ the learned policy deviates from the behavior policy of the offline data.
Besides, in foreign states that ACT is not trained on, it may not be able to generate actions that match the desired advantages. Similar dilemma also bothers policy gradient RL algorithms~\cite{trpo, ppo}. It is interesting to apply the idea of trust-region~\cite{trpo} to test-time policy improvement, and we leave it for future research. 

\section{Implementation Details}
\subsection{Architectures and Hyper-parameters of ACT}\label{app:arch_and_params_act}
In this section, we provide a detailed description of the implementations of ACT as well as other baseline methods.
The hyper-parameters for ACT, which are shared across all of the evaluation tasks, are listed in Table~\ref{apptab:hyper-parameters_act}. For hyper-parameters that vary for different tasks, we will elaborate on them in the corresponding subsections.  It is worth noting that for most of the parameters, we directly inherit their value from existing literature without tuning. Our code is available at 

\begin{table}[htbp]
\centering
\begin{normalsize}
\begin{tabular}{c|c|c}
\toprule[1.5pt]
\textbf{Architecture} &\textbf{Hyper-parameters} & \textbf{Value}\\
\midrule
\multirow{6}{*}{$Q_\theta$ and $V_\psi$} & \# of hidden layers & 3\\
&Layer width & 256\\
&Learning rate &$3\times 10^{-4}$\\
&Discount $\gamma$ & 0.99\\
&Polyak coeff. & 0.005\\
&Batch size & 256\\
&Training steps & 200k\\
&Expectile $\sigma_1$ & 0.5 or 0.7\\
\midrule
\multirow{5}{*}{$c_\phi$} & \# of hidden layers & 3\\
& Layer width & 256\\
&Learning rate & $3\times 10^{-4}$\\
&Batch size & $256$\\
&Training steps & 500k\\
&Expectile $\sigma_2$ & 0.98\\
&Weight decay & $5\times 10^{-4}$\\
\midrule
\multirow{14}{*}{ACT}&\# of encoder layers & 3\\
&\# of decoder layers & 3\\
&\# of attention heads & 1\\
&Embedding dimension & 128 \\
&Nonlinearity function & ReLU \\
&Batch size & 64\\
&Context length & 20 \\
&Dropout & 0.1 \\
&Learning rate & $10^{-4}$\\
&Lr Warmup steps & 10000\\
&Grad norm clip & 0.25 \\
&Weight decay & $10^{-4}$ \\
&Adam betas & [0.9, 0.999]\\
&Training steps & 500k or 1M\\
\bottomrule[1.5pt]
\end{tabular}
\end{normalsize}
\caption{Hyper-parameters of ACT}
\label{apptab:hyper-parameters_act}
\end{table}

\textbf{Value functions. }We parameterize the value functions $Q_\theta$ and $V_\psi$ as 3-layer MLPs respectively, and use Equation~\ref{eq:dp} to iteratively optimize the value functions. To overcome the over-estimation issue~\cite{td3}, we maintain two independent value networks $\hat{V}_{\psi_1}$ and $\hat{V}_{\psi_2}$ and use the minimum of their outputs for bootstrapping. We also introduce target networks $\hat{V}_{\bar{\psi}_1}, \hat{V}_{\bar{\psi}_2}$ to give target values for stability. The parameter $\bar{\psi}_i$ is updated in a moving average manner, i.e., $\bar{\psi}_i^{k+1} = (1-\alpha)\bar{\psi}_i^{k} + \alpha \psi_i^{k+1}$, with the coefficient $\alpha$ being $0.005$ across all tasks. 

\textbf{Predictor network $c_\phi$. }The condition network is also parameterized as a 3-layer MLP, and optimized toward the $0.98$ expectile of the advantage.

\textbf{ACT. }The encoder part of ACT consists of three attention layers, each of which is composed of a self-attention module and a feed-forward network. The attention part uses the causal mask to mask out future tokens to prevent information leaks. For positional encoding, we use sinusoidal positional encoding which is proposed in the original Transformer~\cite{attention_is_all_you_need}. 

The decoder part is adapted from the Transformer decoder. Specifically, we feed a sequence of target advantages into the decoder layer and apply 1) a diagonal mask on the advantage sequence, so that advantages at other timesteps are masked out; and 2) a causal mask on the output sequence of the encoder, so as to prevent information leak from the future. In this way, we are able to implement the core component of ACT within the transformer architecture. 

Finally, we further project the output from the decoder with a linear layer and obtain the predicted deterministic actions. 

\subsection{Details about Experiments on Deterministic Gym MuJoCo tasks}\label{app:impl_deterministic}
We provide details about ACT and the baseline methods involved in our evaluation. 

\textbf{Environments and datasets. }We use Gym MuJoCo, which includes a wide range of deterministic and continuous control tasks to assess the performance of ACT as well as the baseline methods. For the offline dataset, we choose three levels of quality: 1) \textit{medium}, which is collected by a pre-train SAC~\cite{sac} policy that achieves 30\% of the expert performance; 2) \textit{medium-replay}, which consists of the data in the replay buffer of a medium-level policy; 3) \textit{medium-expert}, which is a half-half mixture of data collected by a medium-level policy and an expert-level policy. We use the \textit{-v2} version of datasets. 

\textbf{ACT. }On this deterministic benchmark, we sweep the value of $\sigma_1$ in $\{0.5, 0.7\}$ and use GAE with $\lambda=0$. GAE($0$) has lower bias compared to IAE since it gets rid of the approximated state-action value function $\hat{Q}_\theta$ while not introducing additional variance due to determinism of the environment.
During the course of training, we found that the inherent instability of TD training can lead to occasional oscillations in the value function. Therefore, for this benchmark, we additionally introduce a stage of model selection to assist us in identifying well-fitted models. Specifically, we train in parallel a group of value functions $\{\hat{V}_{\psi_i}\}_{i=1}^n$, each of which is trained with distinct data batches and bootstraps independently.
After the training is done, we use the one with the lowest TD error on the whole offline dataset, measured by $\frac 1{|\mathcal{D}|}\sum_j\mathcal{L}_{\sigma_1}(r_j+\gamma \hat{V}_{\psi_{i}}(s_j') - \hat{V}_{\psi_i}(s_j))$, to label the offline dataset and proceed with subsequent steps.
The rationale behind this is that lower error on the offline dataset means better alignment with the observed rewards. Recent work~\cite{regulation} also reveals that, lower TD error on the validation set is a positive signal for higher performance.
In our setting, the validation/test set is exactly the offline dataset since we are actually using the value functions to label the offline datasets. 

The choice of $\sigma_1$ for each task is listed in Table~\ref{apptab:choice_of_sigma1}. For most of the datasets, $\sigma_1=0.7$ works the best since it provides action evaluation from the perspective of an improved policy. For some tasks, $\sigma_1=0.5$ works the best, probably due to the reasons discussed in Appendix~\ref{app:test-time_policy_improvement}. 

\begin{table}[htbp]
    \centering
    \begin{tabular}{c|c}
    \toprule[1.0pt]
    Dataset & $\sigma_1$\\
    \midrule
     halfcheetah-medium  &  0.7\\
     halfcheetah-medium-replay & 0.7\\
     halfcheetah-medium-expert & 0.7\\
     hopper-medium & 0.5\\
     hopper-medium-replay & 0.7\\
     hopper-medium-expert & 0.5\\
     walker2d-medium & 0.5\\
     walker2d-medium-replay & 0.5\\
     walker2d-medium-expert & 0.7\\
    \bottomrule[1.0pt]
    \end{tabular}
    \caption{Choices of $\sigma_1$ for each dataset. }
    \label{apptab:choice_of_sigma1}
\end{table}

\textbf{Decision Transformer. }To make a fair comparison, we re-implemented DT according to the code provided at the official repository and modified the transformer layers of DT to $6$ to match the parameter count of ACT. For the initial RTG during test time, we reused the RTGs in the authors' choice, i.e., $\{6000, 12000\}$ for \textit{halfcheetah}, $\{1800, 3600\}$ for \textit{hopper}, and $\{2500, 5000\}$ for \textit{walker2d}, and selected the higher score as the result. 

\subsection{Details about the 2048 Game}\label{app:impl_2048}
\textbf{Environment and datasets. }The 2048 game is a $4\times 4$ board game. In the beginning, some of the grids are populated with randomly placed 2 or 4. Each time the player can choose a direction and the tiles on the board are moved along the direction. The tiles with identical values are combined into one tile with double the values. The goal of this environment is to reach a tile of 128. The episode will be ended when the tile of 128 is produced and a reward of 1 will be correspondingly assigned. Thus, the maximal possible return is exactly 1. We use the existing implementation and dataset of this game from~\cite{esper}. The dataset consists of 5M steps of data which are collected by a mixture of a random policy and an expert policy trained with PPO~\cite{ppo}.

\begin{figure}[htbp]
    \centering
    \includegraphics[width=0.8\linewidth]{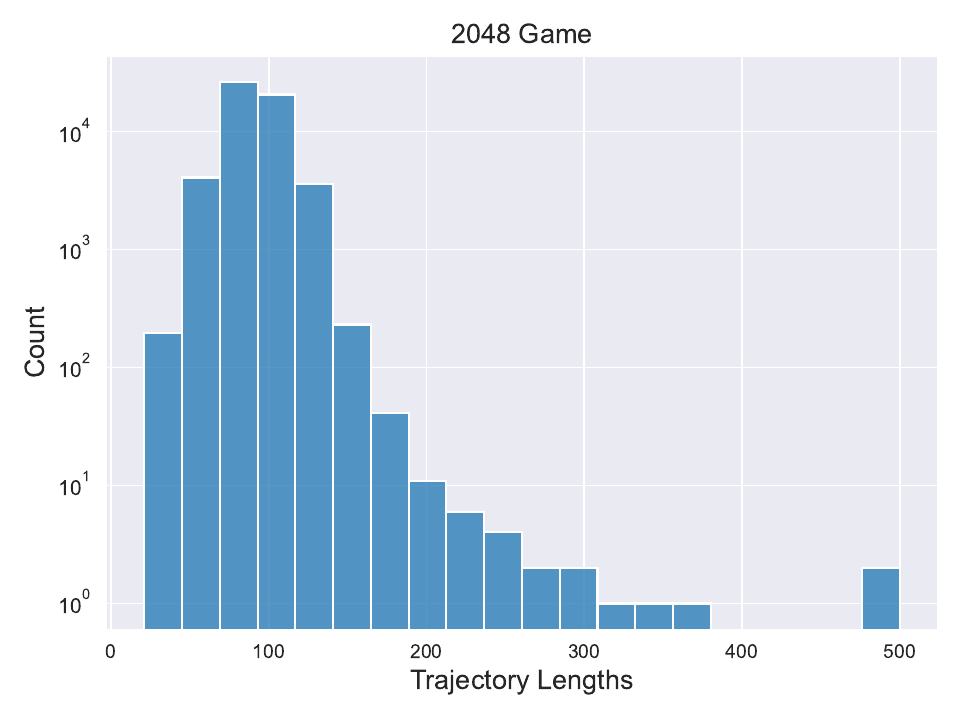}
    \caption{Distribution of the trajectory lengths in the 2048 datasets.}
    \label{appfig:2048_lengths_dist}
\end{figure}

We list the distribution of trajectory lengths and returns in Figure~\ref{appfig:2048_lengths_dist} and Table~\ref{apptab:2048_returns_dist}.

\begin{table}[htbp]
    \centering
    \begin{tabular}{c|c}
    \toprule[1.0pt]
    Return & \# of trajectories\\
    \midrule
    0.0 & 25741\\
    1.0 & 29085\\
    \midrule
    Sum & 54826\\
    \bottomrule[1.0pt]
    \end{tabular}
    \caption{Distribution of the returns in the 2048 dataset. The return for each trajectory is either 0 or 1, so we list the counts for each type as well as the total number of the trajectories.}
    \label{apptab:2048_returns_dist}
\end{table}

\textbf{ACT. }In this experiment, we use IAE to account for the stochasticity of the environment when assessing the actions. For $\sigma_1$, it is set to $0.5$, because $\sigma_1=0.5$ means we are improving the policy on the basis of the behavior policy, thus isolating the benefits that come from trajectory stitching and the consideration of stochasticity. Moreover, we take the average rather than the minimum of the double V-value functions since the over-estimation issue is not severe as we are performing on-policy value estimation. 

Finally, this game is a discrete control task, so we modified the output layer of ACT to predict the logits of a categorical distribution. The training objective for ACT correspondingly transforms to maximizing the log-likelihood of the ground-truth action. 

\textbf{Decision Transformer. }We make a similar modification to DT to adapt it for discrete control. For 2048 game, we set the initial RTG during test time as $1.0$, since $1.0$ is a known upper bound of the returns in this environment. 

\subsection{Details about the Stochastic Gym MuJoCo Tasks}\label{app:impl_stochastic}
\textbf{Environments and datasets. }The original Gym MuJoCo tasks are deterministic both in terms of state transition and reward mechanism. To fulfill the stochasticity, we follow the practice of \citet{doc} to add noise to the action chosen by the policy before it is passed to the simulator. Thus from the perspective of the agent, the outcome from the environment becomes unpredictable, which simulates stochasticity. The noise added to the action is a zero-meaned Gaussian noise $\epsilon\sim\mathcal{N}(0, 1-\exp(-0.01t)\cdot \omega\sin(t))$. For all tasks, $\omega$ is set to $0.1$. We train a SAC policy in the stochastic version of the environments for $300$k steps and log the replay buffer as the offline datasets. 

\begin{figure}[htbp]
    \centering
    \includegraphics[width=1.0\linewidth]{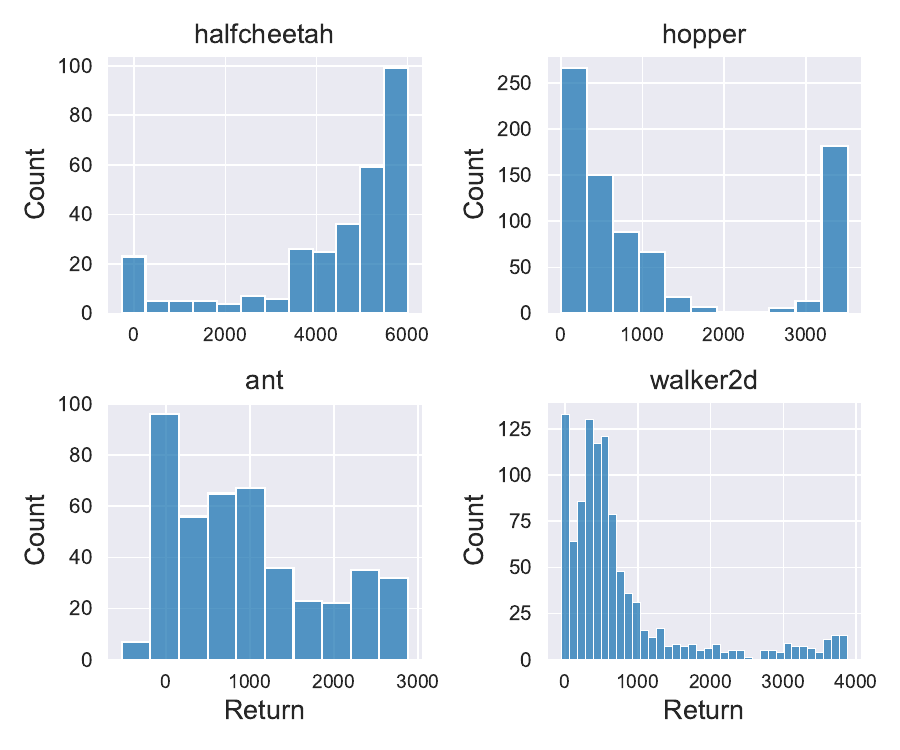}
    \caption{Distributions of the returns in the stochastic Gym MuJoCo datasets.}
    \label{appfig:stoc_gym_return_dist}
\end{figure}

The distributions of the trajectory returns in offline datasets are plotted in Figure~\ref{appfig:stoc_gym_return_dist}. 

\textbf{ACT. }The configurations of ACT are the same as those in the 2048 game, so we will not get into the details again here. It is worthwhile to note that although setting $\sigma_1$ to $0.7$ rather than $0.5$ may further boost the performance, we avoided doing this for the purpose of isolating the benefits that come from stitching trajectories and accounting for the stochasticity. 

\textbf{Decision Transformer. }
The configurations of DT are mostly the same as those used in the deterministic version of the environments. The only difference lies in the initial RTG. We set the RTG according to Table~\ref{apptab:choice_of_RTG}.

\begin{table}[htbp]
    \centering
    \begin{tabular}{c|c}
    \toprule[1.0pt]
    Dataset & Initial RTG $\hat{R}_0$\\
    \midrule
     halfcheetah-stochastic  &  $\{6000, 12000\}$\\
     hopper-stochastic & $\{1800, 3600\}$\\
     walker2d-stochastic & $\{2500, 5000\}$\\
     ant-stochastic & $\{2000, 4000\}$\\
    \bottomrule[1.0pt]
    \end{tabular}
    \caption{Choices of $\hat{R}_0$ for stochastic Gym MuJoCo tasks. }
    \label{apptab:choice_of_RTG}
\end{table}

For each task, we designate two initial RTGs, whose values are determined based on the distribution of returns in the offline datasets. We report the higher score between the RTG candidates during evaluation. 

\begin{figure*}[htbp]
    \centering
    \includegraphics[width=0.9\linewidth]{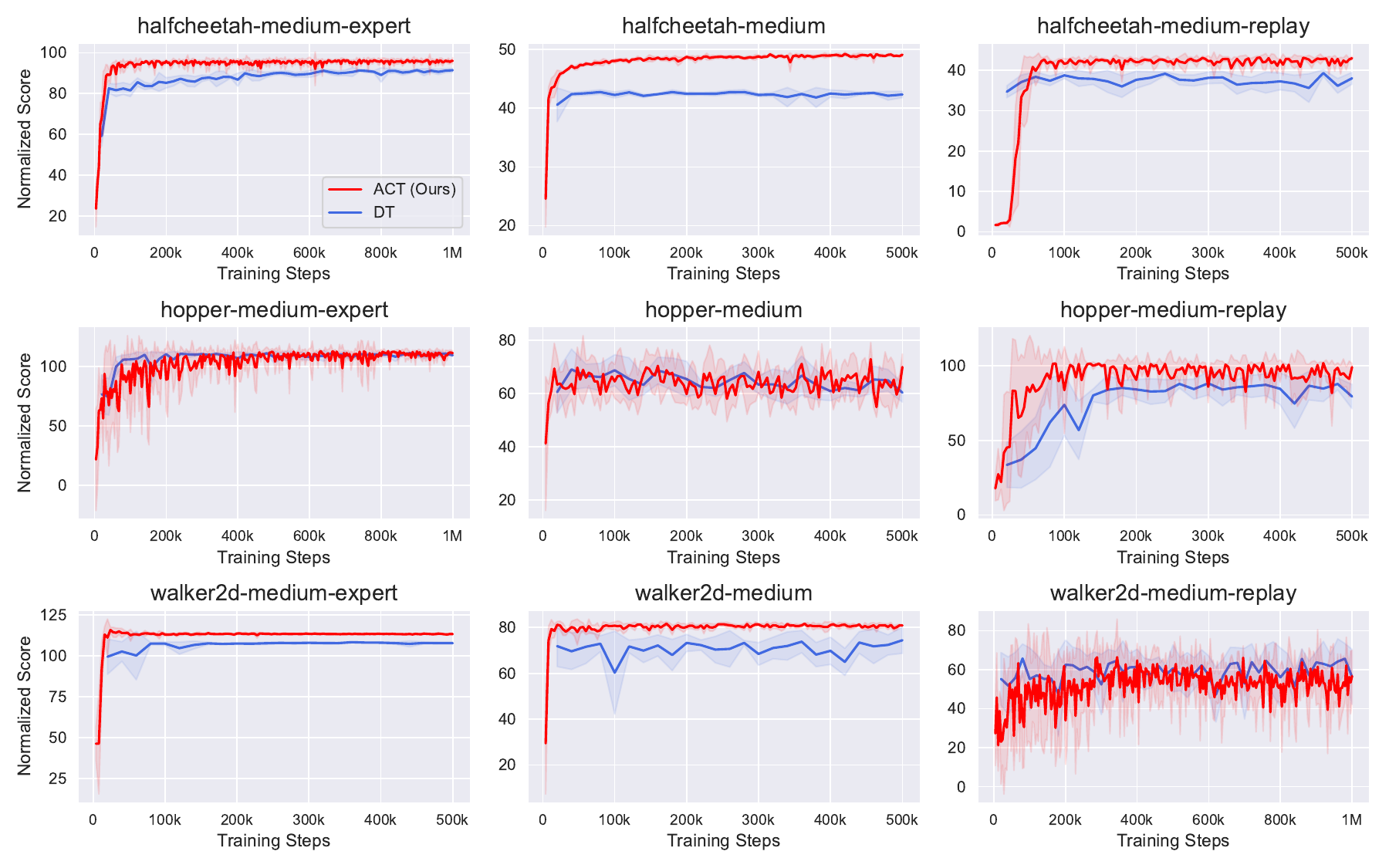}
    \caption{Training curves of DT and ACT on the D4RL datasets. We report the average and one standard deviation across 5 independent runs.}
    \label{appfig:act_curves}
\end{figure*}

\subsection{Details about the Delayed Reward Tasks}

\textbf{Environments and datasets. }For this part, we select \textit{hopper} and \textit{walker2d} tasks from Gym MuJoCo, and the \textit{medium} and \textit{medium-expert} datasets from D4RL as the offline datasets. The reason that we omit \textit{halfcheetah} and \textit{medium-replay} is that the datasets may contain timeout trajectories or fragments of the trajectories, which damages the value estimation since the reward is not always given at the terminal states. To simulate the delayed reward setting, we modify the reward at the last timestep to the cumulative rewards of this trajectory and set the past rewards to zeros. 

\textbf{CQL and IQL. }We use the implementation from CORL~\cite{corl} for IQL and CQL. 

\textbf{Decision Transformer. }
For DT, we use our own implementation which shares a similar parameter count to our ACT. The initial RTGs are kept the same as the original implementation, as in Table~\ref{apptab:choice_of_RTG_delayed}. 

\textbf{ACT. }For ACT, we set $\sigma_1=0.5$, $\gamma=1.0$, and GAE with $\lambda=1.0$. Moreover, we take the average rather than the minimum of the double V-value functions when estimating the values. Other hyper-parameters are kept the same as in Table~\ref{apptab:hyper-parameters_act}.

\begin{table}[htbp]
    \centering
    \begin{tabular}{c|c}
    \toprule[1.0pt]
    Dataset & Initial RTG $\hat{R}_0$\\
    \midrule
     hopper & $\{1800, 3600\}$\\
     walker2d & $\{2500, 5000\}$\\
    \bottomrule[1.0pt]
    \end{tabular}
    \caption{Choices of $\hat{R}_0$ for delayed reward tasks. }
    \label{apptab:choice_of_RTG_delayed}
\end{table}

\section{Supplementary Results and Analysis}
\subsection{Training Curves}
In this section, we provide the performance curves of ACT and DT on the D4RL datasets to present the actual dynamics of training. The results are illustrated in Figure~\ref{appfig:act_curves}. 

\subsection{Ablations on Effect of $\sigma_2$}\label{app:more_perf_abl_sigma2}
\begin{figure*}[htbp]
    \centering
    \includegraphics[width=0.95\linewidth]{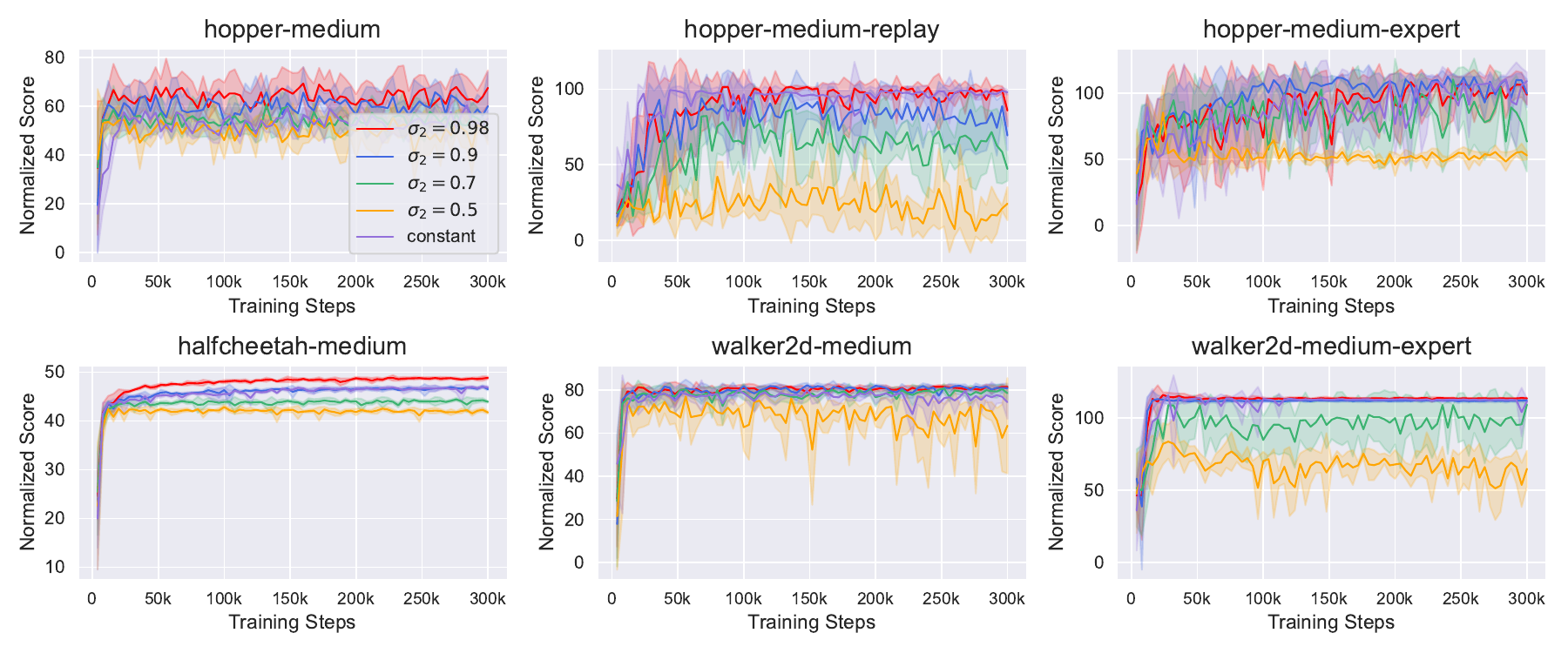}
    \caption{Training curves of ACT with different values of $\sigma_2$ and the constant target advantage. We report the average and one standard deviation across 5 independent runs.}
    \label{appfig:sigma2_curves}
\end{figure*}

In Section~\ref{sec6}, we conducted an ablation study on the effect of $\sigma_2$. In this section, we provide an extended comparison of ACT variants with different values of $\sigma_2$, and the results are illustrated in Figure~\ref{appfig:sigma2_curves}. The trend that higher $\sigma_2$ leads to improved performance is consistent on all datasets. 

\subsection{Ablations on Transformer Architecture}\label{app:more_perf_abl_arch}
\begin{figure*}[htbp]
    \centering
    \includegraphics[width=0.95\linewidth]{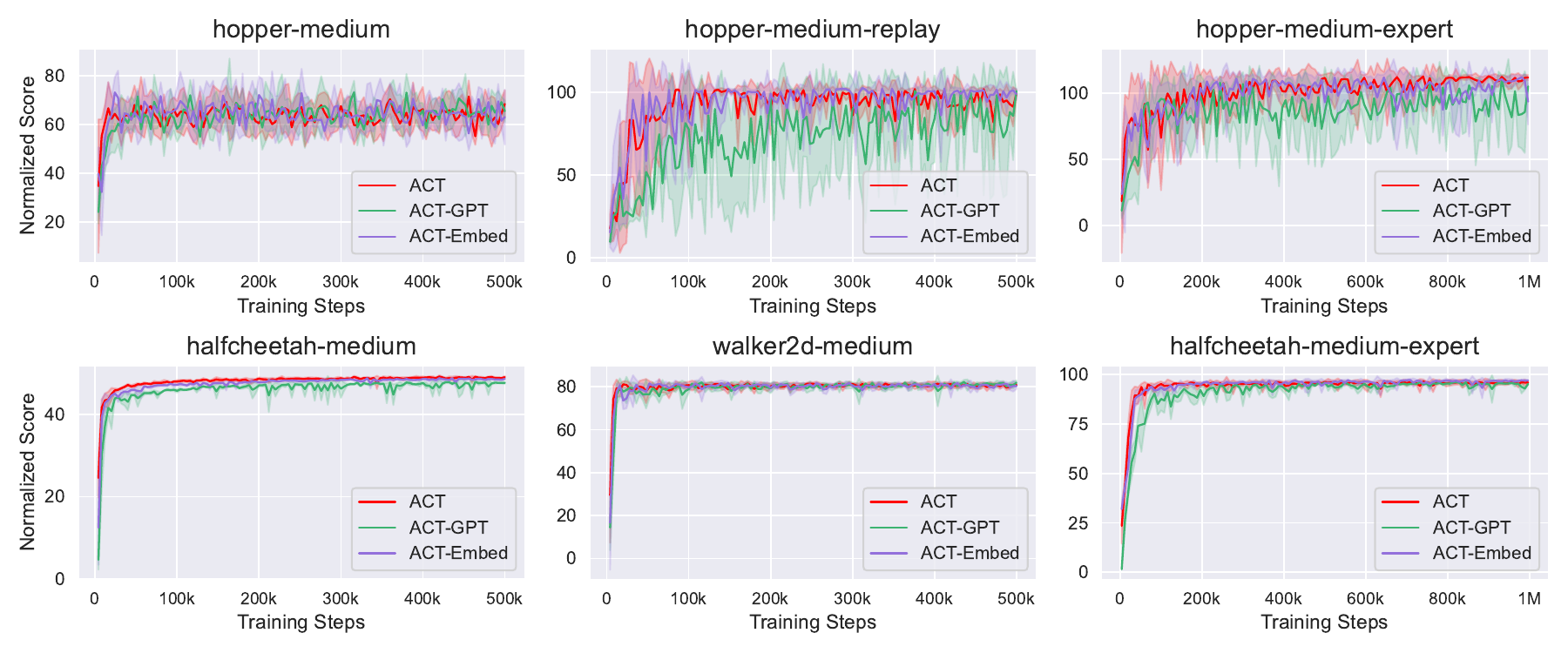}
    \caption{Training curves of ACT with different architectural designs. We report the average and one standard deviation across 5 independent runs.}
    \label{appfig:arch_curves}
\end{figure*}

Another ablation study focuses on transformer architecture, and in this section, we also provide a detailed comparison with more datasets from D4RL. The results are illustrated in Figure~\ref{appfig:arch_curves}. On several datasets \texttt{ACT-GPT} suffers from oscillation and inferior performance. Overall the gap between sinusoidal encoding and the learnable positional embedding is minor. 

\begin{figure}[htbp]
    \centering
    \includegraphics[width=1.0\linewidth]{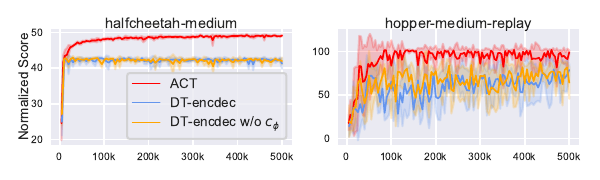}
    \caption{Performances of DT variants using the ACT architecture. The results are taken from 4 independent runs.}
    \label{appfig:abl_arch_dt}
\end{figure}

Moreover, to isolate the effect of advantage conditioning from the effect of architectural design, we also created a variant, \texttt{DT-encdec}, by using the encoder-decoder structure of ACT but keeping the RTG conditioning. We also built another variant based on \texttt{DT-encdec} which removes the need for a predictor $c_\phi$ and calculates the desired RTG as in DT. Their performances are illustrated in Figure~\ref{appfig:abl_arch_dt}. ACT still outperforms the DT variants by a large margin, indicating that both advantage conditioning and the encoder-decoder structure offer benefits. 

\subsection{Running Time Analysis}
Throughout the experiments, we evaluate ACT as well as other baseline methods on workstations equipped with NVIDIA RTX A6000 cards. The running time of each method for D4RL datasets is listed in Table~\ref{apptab:running_time}. 

\begin{table}[htbp]
    \centering
    \begin{tabular}{c|c}
    \toprule[1.0pt]
    Algorithm & Running Time\\
    \midrule
     IQL & 33min\\
     CQL & 2h 51min\\
     DT & 1h 04min\\
     ACT & 1h 49min\\
    \bottomrule[1.0pt]
    \end{tabular}
    \caption{Running time of different methods. }
    \label{apptab:running_time}
\end{table}

\begin{figure}[htbp]
    \centering
    \includegraphics[width=1.0\linewidth]{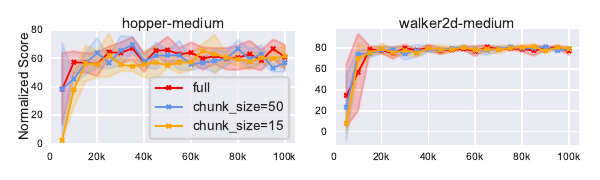}
    \caption{ACT trained with data of different chunk sizes. The results are taken from 4 independent runs.}
    \label{appfig:re_chunk}
\end{figure}

\subsection{ACT with Broken Segments of Data}
In Section~\ref{sec3}, we demonstrate that the integrity of offline trajectories has a significant impact on the performance of DT. In this section, we aim to investigate whether it also affects ACT. 

Theoretically, in ACT, we use the value function $V_\psi(s_t)$ to estimate the expected discounted cumulative reward starting at $s_t$. During training, we only require the successor state $s_{t+1}$ for bootstrapping. When evaluating $a_t$ at $s_t$, we do not require subsequence trajectory if using IAE and only need $s_{t+1}$ if using GAE($0$). On the contrary, calculating RTGs demands the complete subsequent trajectory. We also tested ACT with broken segments of offline data, and the results in Figure~\ref{appfig:re_chunk} do support our claim. 

\subsection{Results on NeoRL}
We also compared ACT to baseline methods on the NeoRL benchmark~\cite{neorl}, a near real-world benchmark for offline RL. This benchmark bears distinct characteristics in that 1) the offline datasets in NeoRL are collected by conservative behavior policies rather than exploratory policies in D4RL, and 2) NeoRL provides validation datasets for model selection. These make NeoRL more compliant with real-world scenarios. 

We select three tasks $\{\textit{HalfCheetah}, \textit{Hopper}, \textit{Walker2d}\}$ and three dataset qualities $\{\textit{low}, \textit{medium}, \textit{high}\}$ from NeoRL, and the results are listed in Table~\ref{tab:neorl_perf}.

Overall, ACT's performance is on par with IQL and DT. However, in this particular benchmark, CQL's performance was observed to be significantly superior to other methods. This can be attributed to the unique characteristics of the NeoRL benchmark. Unlike IQL, DT, and ACT, which rely solely on samples from the offline dataset for policy evaluation and improvement, CQL instead searches along the Q-value function for optimal actions. Although this introduces risks of overestimation, it also permits CQL to take usage of the generalization ability of Q-value functions to escape the limitation of the offline dataset. We conjecture that this is vital for the NeoRL benchmark. Consequently, it may be beneficial to incorporate learned advantage functions in ACT for out-of-distribution (OOD) action generalization.

\begin{table}[htbp]
    \centering
    \scalebox{0.93}{
    \begin{small}
    \begin{tabular}{c|lllll}
    \toprule[1.0pt]
        Dataset & BC & CQL & IQL & DT & ACT\\
        \midrule
        hc-low & $ 29.1 $ & $38.2$ & $ 30.6_{\pm0.3} $ & $ 28.9_{\pm0.1} $ & $ 29.5_{\pm0.3} $\\
        hc-med & $ 49.0 $ & $54.6$ & $ 50.9_{\pm0.8} $ & $ 52.3_{\pm0.2} $ & $ 51.4_{\pm0.2} $\\
        hc-high & $ 71.3 $ & $77.4$ & $ 75.4_{\pm0.1} $ & $ 74.6_{\pm0.1} $ & $ 75.4_{\pm0.6} $\\
        ho-low & $ 15.1 $ & $16.0$ & $ 16.2_{\pm0.2} $ & $ 15.3_{\pm0.4} $ & $ 14.2_{\pm0.6} $\\
        ho-med & $ 51.3 $ & $64.5$ & $ 58.4_{\pm18.8} $ & $ 56.5_{\pm5.1} $ & $ 54.5_{\pm15.5} $\\
        ho-high & $ 43.1 $ & $76.7$ & $ 26.9_{\pm1.0} $ & $ 59.8_{\pm15.2} $ & $ 57.9_{\pm25.0} $\\
        wk-low & $ 28.5 $ & $44.7$ & $ 42.4_{\pm0.6} $ & $ 35.1_{\pm3.9} $ & $ 30.1_{\pm0.7} $\\
        wk-med & $ 48.7 $ & $57.3$ & $ 61.3_{\pm0.2} $ & $ 46.0_{\pm2.6} $ & $ 63.8_{\pm1.2} $\\
        wk-high & $ 72.6 $ & $75.3$ & $ 74.5_{\pm0.3} $ & $ 70.6_{\pm2.8} $ & $ 74.5_{\pm1.4} $\\
        \midrule
        \bottomrule[1.0pt]
    \end{tabular}
    \end{small}
    }
    \caption{Normalized score on NeoRL tasks. The average and std are taken across 4 independent runs.}
    \label{tab:neorl_perf}
\end{table}

\section{Additional Related Topics}
\textbf{Offline Reinforcement Learning. }Unlike online RL which alternates between environmental interaction and policy optimization, offline RL relies on a pre-collected static dataset to optimize the policy. This method is not burdened by the intensive and costly interactions characteristic of online RL. However, it faces a challenge due to the distribution shift between the offline dataset and the evolving policy. This shift implies that as policy improvement progresses, the dataset becomes less capable of supporting further optimization. To address this issue, existing practices can be categorized as follows. The first category is policy-constraint methods~\cite{bcq,bear,td3bc,prdc}, which regularize the policy from producing Out-Of-Distribution (OOD) actions during the improvement step. Apart from this, we can instead regulate the Q-value functions to avoid exploiting OOD actions by applying conservative penalties, as demonstrated in CQL~\cite{cql}, EDAC~\cite{edac}, and PBRL~\cite{pbrl}. Another line of research~\cite{iql,inac,xql,iac} adopts in-sample learning, which iterates without the usage of counter-factual queries of OOD samples, thereby offering safer policy optimization compared to the above methods. Lastly, model-based offline RL methods~\cite{mopo,combo,mobile} introduce learned dynamics models that generate synthetic experiences to alleviate data constraints, thereby outperforming model-free approaches in certain respects.

\textbf{Reinforcement Learning via Supervised Learning. }The concept of addressing reinforcement learning challenges through supervised learning was originally introduced by Upside Down Reinforcement Learning (UDRL)~\cite{udrl}. Instead of predicting reward-related quantities (such as value functions), UDRL takes these quantities as input and maps them to actions directly. Similarly, RCP~\cite{rcp} enhances the policy network by incorporating elements that specify the desired characteristics of actions, such as TD($\lambda$) return or advantages. On the architectural design, Decision Transformer~\cite{dt} extends RCP by employing expressive sequence models as the policy, while RvS~\cite{rvs} finds out that a two-layer feedforward MLP is sufficient provided the capacity and condition quantity are judiciously selected. As an extension of DT, ACT also addresses offline reinforcement learning through supervised learning. Notably, ACT distinguishes itself as the first to integrate advantage functions into transformer-based sequence models.

\section{An Illustrative Example: Cliff-Walking}
\begin{figure}[htbp]
    \centering
    \includegraphics[width=0.6\linewidth]{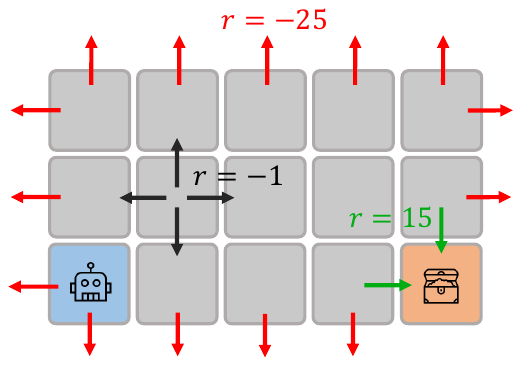}
    \caption{Illustration of the \textit{CliffWalking} environment. }
    \label{appfig:cliffwalking}
\end{figure}

In this section, we introduce a simplified grid-world environment as a probe to illustrate the actual effect of advantage conditioning. The environment is called \textit{CliffWalking}, as shown in Figure~\ref{appfig:cliffwalking}. 

In this environment, the agent is placed at the lower left corner $[0, 0]$ in the beginning. At each timestep, the agent can choose one direction among $[\texttt{Left, Right, Up, Down}]$. If the agent chooses \texttt{Left} or \texttt{Right}, there is a probability of $0.4$ that it actually goes up or down, while for \texttt{Up} and \texttt{Down}, the action will remain unchanged and will be executed faithfully. If the agent reaches the treasure grid placed at the lower right corner $[4, 0]$, it will receive a reward of $10$. If the agent transits out of the grids~(falls off the cliff), it will receive a severe punishment of $-25$. For other scenarios, it will receive a reward of $-1$ as the cost of the move. Each trajectory will be terminated when the agent reaches the goal state, falls off the cliff, or after $30$ moves. The offline dataset, which consists of 10k trajectories, is collected by a random policy. 

To reach the grid with treasures, although walking along the lowest row requires the least steps, it also introduces the risk of falling off the cliff. Thus, it is trivial that the optimal strategy in this environment is to first go upward to $[0, 1]$, then walk through the cliff along the middle row to $[4, 1]$, and finally go down to $[4, 0]$. Whenever the optimal agent deviates from the middle row due to the stochasticity of the environment, it should get back to the middle row as soon as possible. 

Now, we continue to investigate the policy given by DT and ACT. 

\begin{figure}[htbp]
    \centering
    \includegraphics[width=0.8\linewidth]{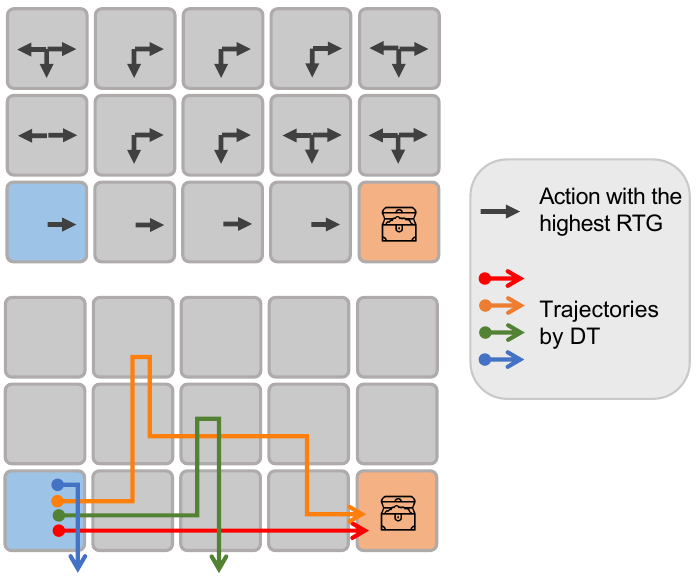}
    \caption{Top: Actions with the highest RTG. Down: Trajectories produced by the trained DT.  }
    \label{appfig:cliffwalking_dt}
\end{figure}

\textbf{Decision Transformer. }
As a beginning step, we calculate the RTGs for data in the offline dataset and mark the action with the highest RTG in Figure~\ref{appfig:cliffwalking_dt}. As we can see, RTGs assess the actions by the most successful past experience, i.e., the shortest path toward the goal state. As a result, the final policy produced by DT keeps choosing \texttt{Right}~(Figure~\ref{appfig:cliffwalking_dt}) to head for the goal state, although such a decision is risky. 

\begin{figure}[htbp]
    \centering
    \includegraphics[width=0.8\linewidth]{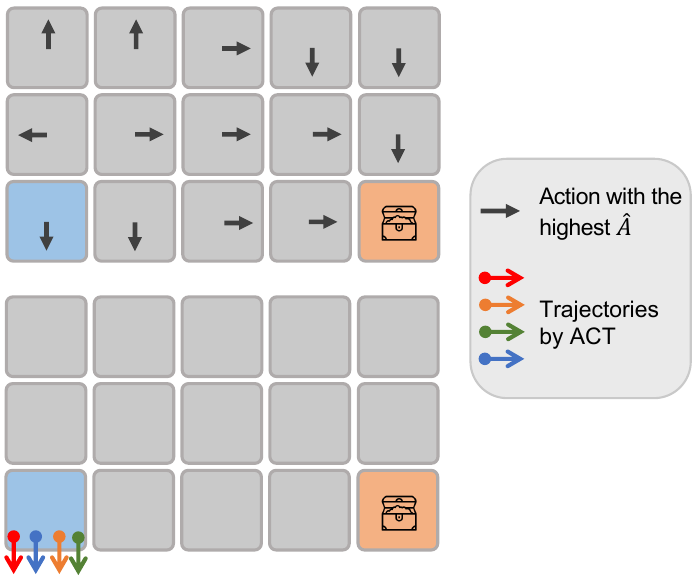}
    \caption{Top: Actions with the highest advantage. Down: Trajectories produced by the trained ACT. The expectile $\sigma_1$ is $0.5$.  }
    \label{appfig:cliffwalking_act0.5}
\end{figure}

\textbf{ACT with $\bm{\sigma_1=0.5}$. }We proceed to investigate the behavior of ACT in this environment. We set $\sigma_1=0.5$ and use IAE as the advantage estimator, which means we are using the estimation of on-policy advantage $\hat{A}^\beta$ to assess the actions. Figure~\ref{appfig:cliffwalking_act0.5} depicts the action with the highest advantage in each state and the final trajectories produced by ACT. We find that when $\sigma_1=0.5$, ACT fails to produce a valid policy, because the dataset is collected by a random policy and the estimated advantage cannot adequately assess the quality of the actions. 

\begin{figure}[htbp]
    \centering
    \includegraphics[width=0.8\linewidth]{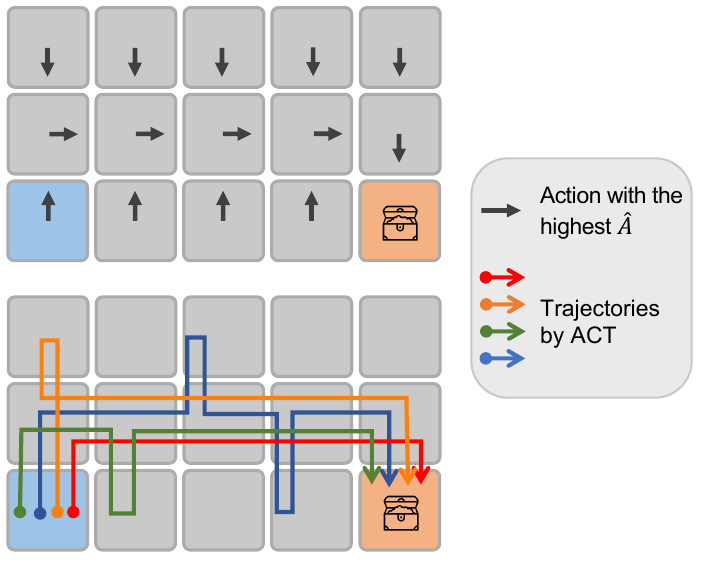}
    \caption{Top: Actions with the highest advantage. Down: Trajectories produced by the trained ACT. The expectile $\sigma_1$ is $0.95$. }
    \label{appfig:cliffwalking_act0.95}
\end{figure}

\textbf{ACT with $\bm{\sigma_1=0.95}$. }Finally, we set $\sigma_1=0.95$ so that we are approximating the advantage of a nearly optimal in-sample policy. Compared to DT and the above variant, this version of ACT uses the advantage of an already improved policy to assess the quality of actions. The results are depicted in Figure~\ref{appfig:cliffwalking_act0.95}. In this version, the advantage suggests the agent first go up to $[0, 1]$, then follow the middle row towards the rightmost column, and finally go down to the goal state. The example trajectories produced by ACT also demonstrate that ACT faithfully executes such optimal strategy. Moreover, whenever it deviates from the middle row, it corrects itself, gets back, and steadily reaches the final goal state in spite of the environmental stochasticity. 

Summarizing the above three experiments, we arrive at two conclusions. First, increasing the value of $\sigma_1$ does bring benefits in some circumstances, as higher $\sigma_1$ allows us to evaluate the actions from the standpoint of an improved policy. Second, compared with RTG conditioning, advantage conditioning gives a robust assessment based on the return in expectation, rather than occasional random success. Such property provides sequence modeling methods with the possibility of converging to the optimal policy, rather than a blindly optimistic policy as return conditioning will do. 
\end{document}